%% file: neurips_2026.tex
\documentclass{article}

 \usepackage[main, final]{neurips_2026}

\usepackage[utf8]{inputenc} 
\usepackage[T1]{fontenc}    
\usepackage{hyperref}       
\usepackage{url}            
\usepackage{booktabs}       
\usepackage{amsfonts}       
\usepackage{nicefrac}       
\usepackage{microtype}      
\usepackage{xcolor}         
\usepackage{adjustbox}
\usepackage{subcaption}
\usepackage{multirow}
\usepackage{wrapfig}
\usepackage{minitoc}

\usepackage{graphicx}
\usepackage{tablefootnote}
\usepackage[table]{xcolor}

\usepackage[most]{tcolorbox} 

\usepackage{algorithmic}
\usepackage{algorithm}
\definecolor{pink}{RGB}{230,150,200}
\definecolor{lightgreen}{RGB}{200,220,140}
\definecolor{purple}{RGB}{210, 170, 230}

\usepackage{amsmath}
\usepackage{amssymb}
\usepackage{mathtools}
\usepackage{amsthm}

\theoremstyle{plain}

\theoremstyle{definition}

\theoremstyle{remark}

\newcommand\mc{\mathcal}

\usepackage[capitalize,noabbrev]{cleveref}

\usepackage[textsize=tiny]{todonotes}

\title{Robust and Efficient Finetuning of Vision Foundation Models via Implicit Ensembling}

\author{
  Masih Aminbeidokhti\textsuperscript{}\thanks{Correspondence to: masih.aminbeidokhti.1@ens.etsmtl.ca. Code at \href{https://github.com/Masseeh/GMixout}{https://github.com/Masseeh/GMixout}.} \quad
  Heitor Rapela Medeiros \quad
  Eric Granger \quad
  Marco Pedersoli \\ \\
  LIVIA, ÉTS Montréal, Canada \\
International Laboratory on Learning Systems (ILLS)
}

\begin{document}

\maketitle

\begin{abstract}
Robust finetuning of foundation models seeks to improve out-of-distribution (OOD) generalization during in-distribution (ID) adaptation. We revisit Mixout, a stochastic regularizer that intermittently replaces finetuned parameters with a reference anchor, and study its effectiveness for robust finetuning of vision foundation models. We reinterpret Mixout as a single-run, weight-sharing implicit ensemble and analyze its expected OOD error through a bias–variance–covariance–locality (BVCL) lens. This perspective reveals three key factors that govern the ID–OOD trade-off: the choice of masking anchor, the mask resampling frequency, and mask sparsity. Guided by this analysis, we propose GMixout, which generalizes Mixout in three ways. First, it explicitly controls the masking period through a resampling-frequency hyperparameter. Second, it replaces the fixed pretrained anchor with an exponential moving average snapshot that adapts throughout training. Third, it uses sparse kernels to update only a small subset of parameters at each iteration, introducing no inference-time overhead and enabling large-model finetuning on consumer-grade GPUs. Experiments across five vision benchmarks show that GMixout consistently outperforms Mixout, Model Soups, and strong parameter-efficient finetuning baselines under both covariate shift and class imbalance.
\end{abstract}

\section{Introduction}

\begin{figure*}[ht]
     \centering
     \includegraphics[width=1.0\textwidth]{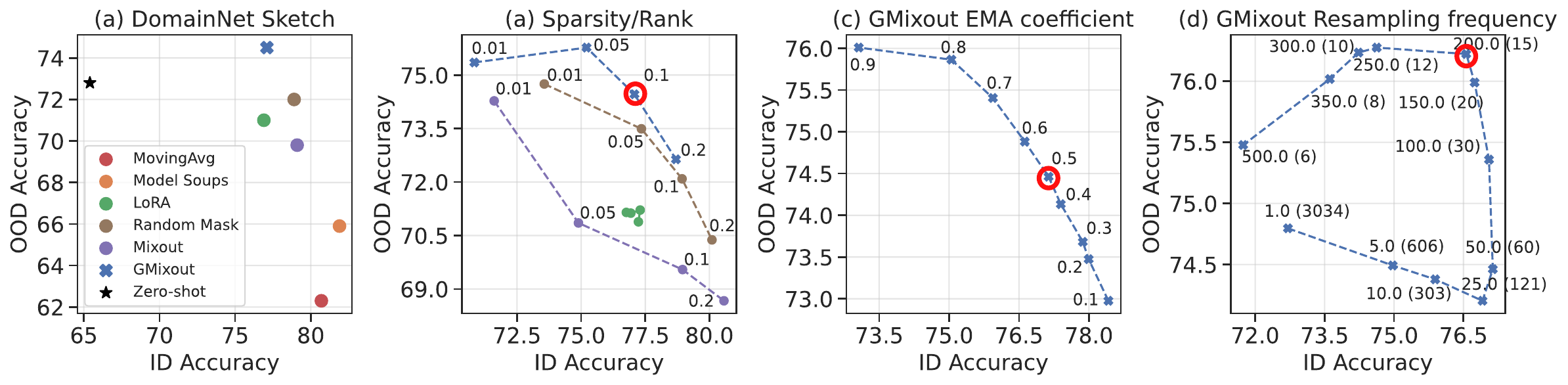}
    \caption{\textbf{OOD-ID accuracy curve on the DomainNet dataset. Models are trained on Sketch (ID) and then evaluated on Real, Painting, and Clipart (OOD) data.} 
(a) Each point represents a method, with ID accuracy on the x-axis and OOD accuracy on the y-axis. The ideal method should improve over the zero-shot baseline along both axes.
(b) Mixout regulates the ID–OOD trade-off through mask sparsity.
(c, d) GMixout extends Mixout with two mechanisms: (i) an EMA coefficient that updates the masking anchor, and (ii) a resampling frequency that determines the number of uncovered subnetworks during optimization (shown in parentheses in (d)). By controlling the variance and covariance of the expected test error, these hyperparameters significantly enhance OOD performance while maintaining competitiveness on ID.}     
\label{fig:intro}
\end{figure*}

A core assumption in standard machine learning is that training and test examples are independently and identically distributed (i.i.d.)~\citep{vapnik1991principles}. In practice, this assumption is often violated: input distributions may shift across environments~\citep{koh2021wilds,gulrajani2020search,yang2023change,mayilvahanan2025in}, models may rely on spurious cues rather than the intended signal~\citep{geirhos2020shortcut}, and class frequencies can be highly imbalanced~\citep{yang2023change,liu2019large}. Foundation models have reshaped learning under distribution shift~\citep{bommasani2021opportunities,radford2021learning,oquab2023dinov2}. Through large-scale and heterogeneous pretraining, they learn transferable representations that can generalize well, even in zero-shot settings~\citep{radford2021learning}. Nevertheless, real-world deployments usually provide some domain-specific data, and finetuning remains the standard way to maximize downstream performance~\citep{tian2023fast,wortsman2022robust}. This makes robust finetuning under distribution shift a central problem.

Existing robust finetuning methods, however, still have important limitations. Domain-invariant representation learning can be overly restrictive and may even underperform strong zero-shot baselines~\citep{arjovsky2019invariant,sun2016deep,ganin2016domain,mayilvahanan2025in}. Augmentation-based methods require invariance-preserving transformations, which often depend on domain expertise and can be difficult or costly to design~\citep{gontijo2020affinity,liu2022empirical,aminbeidokhti2024domain,gao2023out}. Ensembling and Model Soups are among the most reliable approaches for improving robustness, but their training cost scales poorly for large foundation models~\citep{wortsman2022model,rame2022diverse}. Parameter-efficient finetuning (PEFT) methods, such as random masking~\citep{xu2024random} and low-rank adaptation (LoRA)~\citep{hu2022lora}, reduce this cost by updating only a subset of parameters, but their OOD performance can still degrade under even modest shifts from the finetuning distribution~\citep{shuttleworth2024lora,han2024parameter,hu2022lora,xu2024random}.

Implicit-ensemble methods, which interpolate model weights along a single training trajectory, have shown strong performance on OOD benchmarks~\citep{cha2021swad,arpit2022ensemble,izmailov2018averaging}. Yet they are typically based on full-parameter finetuning, making them less practical for large foundation models, especially under limited computational resources. Our preliminary results in Figure~\ref{fig:intro}(a) further show that, when models are finetuned on DomainNet-Sketch and evaluated on Real, Painting, and Clipart, MovingAvg improves in-distribution accuracy but remains substantially weaker than PEFT counterparts on OOD domains. Moreover, none of the considered robust finetuning methods surpasses the zero-shot model under distribution shift. These observations motivate us to revisit implicit ensembling in a parameter-efficient setting, with the goal of combining the practicality of PEFT with the robustness benefits of weight-space averaging.

To this end, we reinterpret Mixout as a parameter-efficient implicit ensemble method. Mixout applies a binary mask during training to randomly replace finetuned weights with their frozen pretrained counterparts, thereby regularizing optimization toward the initialization~\citep{lee2019mixout}. Its sparsity provides a simple mechanism for controlling the trade-off between in-distribution adaptation and robustness. However, although Mixout was effective in the original NLP setting, it is less competitive for vision foundation models. As shown in Figure~\ref{fig:intro}(b), vanilla Mixout does not achieve the same ID--OOD curve as strong PEFT baselines such as LoRA and Random Mask.

We argue that this limitation is better understood from a weight-space ensemble perspective. Under this view, Mixout forms a single-run, weight-sharing implicit ensemble: each binary mask defines a different subnetwork, and training averages over many such subnetworks. Robustness, therefore, depends not only on sparsity, but also on the diversity and correlation among the sampled subnetworks, which determine the variance--covariance behavior of the ensemble~\citep{dietterich2000ensemble,allen-zhu2023towards}. We formalize this intuition using a bias--variance--covariance--locality (BVCL) decomposition~\citep{rame2023model}. This analysis highlights two important design choices that are overlooked in standard Mixout: the anchor weights and the mask resampling frequency. Standard Mixout always reverts to the fixed pretrained initialization and resamples the mask at every step, which can increase subnetwork correlation and bias learning toward an anchor that may no longer be optimal after downstream adaptation.

Guided by this analysis, we propose \underline{G}Mixout, a \underline{generalized} Mixout framework that decouples sparsity, anchor selection, and mask resampling. First, GMixout introduces an explicit resampling-frequency hyperparameter that controls how often weight switching occurs, thereby regulating the diversity and correlation of the implicit subnetworks; see Figure~\ref{fig:intro}(d). Second, GMixout replaces the fixed pretrained anchor with an exponential moving average (EMA) reference that evolves during training, allowing the anchor to track a task-adapted solution while preserving the regularization effect of the pretrained initialization; see Figure~\ref{fig:intro}(c). Finally, we implement GMixout with efficient sparse CUDA kernels~\citep{gale2020sparse}, so that only a small fraction of parameters is updated at each step. The resulting method has no inference-time overhead and enables scalable finetuning of large foundation models on consumer-grade GPUs.

Our contributions are threefold. \textbf{(1)} We reinterpret Mixout as a single-run, weight-sharing implicit ensemble and analyze its ID--OOD behavior through a BVCL framework, identifying anchor weights, mask resampling frequency, and sparsity as key factors controlling the adaptation--robustness trade-off. \textbf{(2)} We introduce GMixout, which generalizes Mixout with an adaptive EMA anchor and an explicit resampling-frequency hyperparameter, together with a memory- and compute-efficient sparse-kernel implementation. \textbf{(3)} We empirically validate GMixout across a broad range of distribution shifts on vision benchmarks, showing consistent improvements over strong PEFT and ensemble baselines while substantially reducing training cost.

\section{Related Work}
\noindent\textbf{Robust finetuning.}
A large body of work studies how to improve the robustness of finetuned foundation models under OOD shifts. In vision--language models, prior studies show that preserving aspects of the pretraining objective can substantially outperform standard cross-entropy finetuning~\citep{goyal2023finetune,oh2024towards}. More broadly, several methods improve robustness by constraining finetuned weights to remain close to the pretrained initialization~\citep{kumar2022finetuning,xuhong2018explicit,tian2023fast,tian2024rethinking}. Another effective direction is ensembling, where methods such as Model Soups aggregate multiple finetuned models to reduce variance and improve robustness~\citep{dietterich2000ensemble,rame2022diverse,allen-zhu2023towards}. However, explicit ensembles require multiple full finetuning runs, making them expensive at foundation-model scale. To reduce the cost of explicit ensembles, prior work has explored single-run alternatives, including weight averaging, mean-teacher updates, robust weight interpolation, and batch ensembles~\citep{izmailov2018averaging,tarvainen2017mean,wortsman2022robust,oh2025dawin,zhu2024robust,wen2020batchensemble}. Although more efficient, these methods often trade off ensemble diversity or still rely on costly full-parameter updates~\citep{bommasani2021opportunities}. GMixout is related to this line of work, particularly adaptive-reference methods such as~\citet{jang2025model}, but realizes a moving-anchor mechanism within a single sparse training run rather than by maintaining or finetuning multiple full-parameter models.

\noindent\textbf{Parameter-efficient finetuning methods.}
Parameter-efficient finetuning (PEFT) reduces adaptation cost by updating only a small subset of parameters or by adding a small number of trainable components~\citep{han2024parameter}. Common PEFT methods include Adapters~\citep{houlsby2019parameter,chen2022adaptformer}, which insert lightweight bottleneck modules into transformer blocks, and Low-Rank Adaptation (LoRA)~\citep{hu2022lora,liu2024dora}, which represents weight updates using low-rank matrices and introduces no inference-time overhead after merging. More recently, \citet{xu2024random} showed that Random Masking, which updates only a fixed, randomly selected subset of parameters, can match or exceed LoRA while retaining the full expressive capacity of the pretrained model. Despite these advances, standard PEFT methods often provide limited OOD robustness gains over zero-shot baselines~\citep{biderman2024lora,shuttleworth2024lora}.

Several recent methods specifically target robust LoRA adaptation. PACE~\citep{ni2024pace} perturbs adapter features with multiplicative noise and enforces consistency across perturbed representations, while SoMA~\citep{yun2025soma} and related approaches~\citep{hu2024learn,meng2024pissa} use SVD-based initialization and selectively tune minor singular components. These methods are competitive on benchmarks such as DomainNet, but we find that they are less effective when the downstream distribution differs more substantially from the pretraining distribution, such as on iWildCam. One possible reason is that the low-rank parameterization can limit adaptation capacity~\citep{hu2022lora}, especially when stronger task-specific adaptation is required.

Prompt tuning is another related PEFT direction, where trainable continuous prompts are added to the model input or intermediate representations~\citep{li2025surveyprompttuning,zhou2022learning}. It is particularly effective for few-shot adaptation in vision--language models~\citep{zhou2022conditional}. In contrast, GMixout is a general finetuning strategy for medium- and large-scale adaptation and is not restricted to the vision--language setting; see Appendix~\ref{appx:prompt-tuning} for further comparisons.

Although GMixout updates only a sparse subset of parameters at each iteration, it exposes the same overall parameter space as full finetuning over the course of training. We therefore view it as a PEFT method under the standard definition, which emphasizes reducing additional parameters, memory, and computational cost rather than strictly limiting the set of accessible parameters~\citep{han2024parameter}.

\noindent\textbf{Dropout.}
Dropout~\citep{srivastava2014dropout} and its variants~\citep{ghiasi2018dropblock,wan2013regularization} are widely used regularization techniques that improve in-domain generalization by randomly deactivating subsets of activations during training. While these techniques are commonly used when training models from scratch, applying dropout directly to pretrained models can disrupt critical learned representations and undermine the benefits of pretraining~\citep{zhang2024fine}. To address this issue, Mixout~\citep{lee2019mixout} randomly replaces finetuned weights with their pretrained values during training, thereby regularizing the model toward the pretrained initialization.

GMixout goes beyond Mixout in four fundamental ways. First, GMixout introduces an explicit resampling-frequency hyperparameter that controls how often weight switching occurs, thereby regulating the diversity and correlation of the implicit subnetworks. Second, whereas Mixout uses a fixed pretrained anchor, GMixout employs a moving anchor that evolves during training, allowing the regularization target to adapt as the model improves. Third, Mixout and most dropout-based methods are designed for full-parameter finetuning, whereas GMixout combines mixout-style regularization with hardware-efficient sparse masked implementation~\citep{gale2020sparse}. This substantially reduces training cost and enables efficient finetuning on consumer-grade GPUs. Finally, Mixout was originally studied in NLP, where it achieved strong results; our preliminary language-model ablations are consistent with these findings. In contrast, this work focuses on vision benchmarks, where we find that vanilla Mixout is less effective than GMixout.

\section{Background}

\noindent\textbf{Task and evaluation.} Given training examples sampled from a distribution $P_{\mathrm{id}}$, our goal is to finetune a foundation model $f : \mathbb{R}^d \rightarrow \mathcal{Y}$ that maps inputs $x \in \mathbb{R}^d$ to outputs $y \in \mathcal{Y}$. We evaluate the model using its standard \emph{in-distribution} (ID) performance on test samples drawn from the same distribution $P_{\mathrm{id}}$. In addition, we assess \emph{out-of-distribution} (OOD) performance on test samples, indicating a different distribution $P_{\mathrm{ood}} \neq P_{\mathrm{id}}$. In practice, $P_{\mathrm{ood}}$ may differ from $P_{\mathrm{id}}$ due to (i) covariate shift in $P(x)$, (ii) class imbalance in $P(y)$, or (iii) changes in $P(y \mid x)$ arising from spurious correlations that are absent at test time. During training, we assume no access to samples from $P_{\mathrm{ood}}$ and therefore minimize empirical risk using only data drawn from $P_{\mathrm{id}}$. For a loss function $\ell$ and a foundation model with parameter vector $\Phi$ this objective is given by:
\begin{equation}
\mathcal{L}(\Phi) \;=\;
\mathbb{E}_{(\boldsymbol{x},y)\sim P_{\text{id}}}
\bigl[\ell\bigl(f(x), y\bigr)\bigr].
\end{equation}

\noindent\textbf{Mixout.} Let $\Phi_0$ denote the frozen pretrained model weights, and let $\Delta$ be learnable parameters of the same shape and initial value as $\Phi_0$. During training with Mixout~\citep{lee2019mixout}, at each iteration, a binary mask is generated, where each element is independently sampled from Bernoulli($1-p$), where $p$ is the masking probability, with the sparsity of $s=1-p$. According to the mask, the corresponding coordinates of the learnable parameter $\Delta$ are then swapped with their pretrained counterparts $\Phi_0$, effectively freezing these coordinates from gradient updates. To ensure that the expected output of layer activations during training matches their output at test time, Mixout rescales $\Delta$ by $(1-p)^{-1}$ during training. This process yields the following risk minimization:
\begin{equation} \label{eq:mixout}
    \min_{\Delta} \mathbb{E}\bigl[\mathcal{L}\bigl( \tfrac{1}{1-p}( (\mathbf{I} - \mathbf{M})\odot\Phi_0 + \mathbf{M}\odot\Delta)\bigr) \bigr],
\end{equation}
where $\mathbf{I}$ is the identity matrix, $\odot$ is the Hadamard product, and the expectation is taken over the random masks $\mathbf{M}$.

\section{Proposed Method}

\subsection{Mixout Error Analysis} \label{sec:gmixout_error}

As shown in~\citet{lee2019mixout}, Mixout can be interpreted as an adaptive form of weight decay, where the effective regularization strength increases with the mask probability $p$, biasing the solution toward the reference parameters $\Phi_0$ (see Appendix~\ref{appx:mixout_wd} for details). While informative, this viewpoint does not capture how the \emph{unmasked subnetworks} interact during training. We therefore develop a complementary perspective based on implicit ensembling in weight space. Consider $M$ finetuned models, each trained under an identically distributed (i.d.) learning procedure $l_{P_{\text{id}}}$. The expected error of a model with weights  $\Phi_{\text{WA}} = \tfrac{1}{M} \sum_{m=1}^M \Phi_m$  with respect to the joint distribution $L_{P_{\text{id}}}^M = \{l_{P_{\text{id}}}^{(m)}\}_{m=1}^M$ can be decomposed into~\citep{rame2022diverse}\footnote{Refer to Appendix~\ref{appx:full_bvcl} for the detailed formulation.}:
\begin{equation} \label{eq:bvcl}
\begin{aligned}
&\mathbb{E}_{L_{P_{\text{id}}}^M}
\bigl[\mc{L}_{\text{ood}}(\Phi_{\text{WA}})\bigr]
=
\mathbb{E}_{(x,y)\sim P_{\text{ood}}}
\Big[
\text{bias}^2(x,y)
+ \frac{1}{M}\,\text{var}(x)
+ \frac{M-1}{M}\,\text{cov}(x)
\Big]
+ O(\bar{\Pi}^2).
\end{aligned}
\end{equation}
Here, the bias is the same as that of
each of its i.d. members. The total variance is split into the variance of each of its i.d. members divided
by $M$ and a covariance term. The locality term $\bar{\Pi}^2$ quantifies the expected squared distance between individual weights and their mean. Eq.~\eqref{eq:bvcl} highlights that averaging $M$ models reduces variance by a factor of $M$, but the resulting OOD error also depends on controlling covariance and locality (in addition to bias). 

\noindent\textbf{Implications for Mixout.}
Now consider the Mixout training procedure. Let $\{\Phi_i\}_{i=1}^I$ denote the sequence of parameters associated with the masked updates over $I$ training iterations, and define the corresponding weight average
$\Phi_{\text{Mixout}} = \tfrac{1}{I}\sum_{i=1}^I \Phi_i$. Since standard Mixout resamples the binary mask at every iteration, it effectively corresponds to an ensemble size $M=I$ in the BVCL view. Combining the ensemble decomposition in Eq.~\ref{eq:bvcl} with the adaptive $\ell_{2}$-penalty view of Mixout reveals the following two key limitations of standard Mixout. \\
\noindent\textbf{(1) Covariance--adaptation trade-off via $p$.}
In standard Mixout, the primary mechanism to reduce covariance across subnetworks is increasing the mask probability $p$, which decreases the overlap between subnetworks and can reduce correlation (Figure~\ref{fig:intro}(b)). However, according to Eq.~\ref{eq:mixout-l2}, a large $p$ also strengthens the implicit pull toward $\Phi_0$, limiting downstream adaptation and making optimization brittle when a small subset of parameters is effectively updated. \\
\noindent\textbf{(2) Fixed anchor can hinder adaptation.}
Mixout always reverts masked parameters to a fixed anchor $\Phi_0$. This is beneficial when $\Phi_0$ is already well-aligned with the downstream task, but can be restrictive when the task requires substantial adaptation (e.g., when zero-shot performance is poor as in iWildCam benchmark). In such cases, a fixed anchor increases bias and prevents the implicit ensemble from tracking a better reference throughout training.

These findings are further supported by the empirical results in Section~\ref{ablations} and motivate three fundamental modifications to Mixout, which we formalize next.

\subsection{From Mixout to GMixout}

GMixout generalizes Mixout along three dimensions that directly address the limitations
identified in the previous section. First, it introduces a resampling-frequency hyperparameter $k$, which controls how often the binary mask is refreshed; standard Mixout is recovered as the special case $k=1$. By decoupling mask refresh from the total training iterations, it provides a direct handle on the diversity--covariance trade-off of sampled subnetworks, without relying solely on increasing $p$. Second, GMixout replaces the fixed pretrained anchor $\Phi_0$ with an adaptive reference
updated via an EMA. Concretely, we partition training into \emph{episodes} indexed by $i$, where each episode consists of $k$ consecutive optimization steps performed under a fixed random mask. Let $\Delta_i$ denote the parameter accumulated within episode $i$. At the end of the episode, the anchor is updated as $\Phi_i = \lambda \Phi_{i-1} + (1-\lambda)\Delta_i$ with $\lambda\in[0,1)$, and the $\Delta$ is reset (i.e., the next episode starts from $\Delta_{i+1} \leftarrow \Phi_i$). This moving anchor allows the implicit ensemble to track a task-adapted reference, mitigating the bias induced by reverting to a static $\Phi_0$ when substantial adaptation is required. The resulting risk minimization objective follows Eq.~\ref{eq:mixout}, with $\Phi_0$ replaced by the episode-dependent anchor $\Phi_i$. In practice, $k$ is chosen as a function of the total training iterations; for clarity, we parameterize it via $E$, the total number of episodes. Finally, unlike the standard Mixout implementation which uses a dense matrix for masking, GMixout keeps $(\mathbf{S}, \Delta_\mathbf{S})$ for the unmasked indices $\mathbf{S}$ and reconstructs $\mathbf{M}\odot\Delta$ with sparse CUDA kernels~\citep{gale2020sparse}. At inference, we merge $\Delta$ into $\Phi_0$, therefore adding no extra cost on the forward pass. See Algorithm~\ref{alg:gmixout} for the GMixout procedure and its differences from standard Mixout.

\begin{algorithm}[ht]
\caption{\colorbox{lightgreen}{GMixout} and \colorbox{pink}{Mixout} Training Procedure}
\label{alg:gmixout}
\begin{algorithmic}[1]
    \STATE Initialize model parameters $\Phi_0$, mask probability $p$, total number of episodes $E$, EMA coefficient $\lambda$, total training iterations $T$. \\
     Define $\Delta \leftarrow \Phi_0$,  \colorbox{pink}{$k \leftarrow 1$} or \colorbox{lightgreen}{$k \leftarrow \lfloor \tfrac{T}{E} \rfloor$}
    \FOR{each iteration $iter$}
        \IF{$iter \bmod k = 0$}
            \STATE \colorbox{lightgreen}{$\Phi_i \leftarrow \lambda \Phi_{i-1} + (1-\lambda)\Delta$}
            \STATE \colorbox{lightgreen}{$\Delta \leftarrow \Phi_i$}
            \STATE $\mathbf{M} \sim \text{Bernoulli}(1-p)$
        \ENDIF
        \STATE Update parameters according to Eq.~\ref{eq:mixout}.
    \ENDFOR
\end{algorithmic}
\end{algorithm}

\section{Results and Discussion} \label{main:results}
\noindent\textbf{Models and Datasets.} 
We use a ViT-B/16~\citep{dosovitskiy2020image} pretrained on OpenAI’s CLIP~\citep{radford2021learning} dataset as our main vision backbone, due to its strong robustness. However, as shown in our ablation studies (Section~\ref{ablations}), GMixout is not tied to this initialization and generalizes across alternative pretrained weights, including both vision-only and language-only models. We evaluate GMixout under two major distribution-shift settings: covariate shift and long-tailed class imbalance. For covariate shift, we consider \texttt{ImageNet-1K}~\citep{russakovsky2015imagenet} with OOD evaluation on \texttt{ImageNet-V2}~\citep{recht2019imagenet}, \texttt{ImageNet-R}~\citep{hendrycks2021many}, \texttt{ImageNet-Sketch}~\citep{wang2019learning}, and \texttt{ImageNet-A}~\citep{hendrycks2021natural}; \texttt{DomainNet}~\citep{peng2019moment}, following prior work~\citep{kumar2022finetuning,tian2024rethinking} by training on one domain and evaluating on the remaining domains, repeated for each source domain; and \texttt{iWildCam}~\citep{beery2021iwildcam} from WILDS~\citep{koh2021wilds}, where we use the standard split with an ID validation set drawn from diverse training environments and an OOD test set from unseen environments. For long-tailed class imbalance, we evaluate on \texttt{CIFAR100-LT}~\citep{cao2019learning} and \texttt{ImageNet-LT}~\citep{liu2019large}. Full dataset descriptions are provided in Appendix~\ref{appx:datasets}.

\noindent\textbf{Baselines.} Across all benchmarks, we compare GMixout against LoRA, Random Masking, Mixout, full-parameter finetuning (Full-FT), its moving average variant (MovingAvg~\citep{arpit2022ensemble}), and Model Soups constructed from five independently finetuned models. These methods represent widely adopted and competitive approaches on robustness-focused vision benchmarks~\citep{rame2022diverse,shi2024long,yang2023change,koh2021wilds}. Additionally, for DomainNet and iWildCam, we include task-specific state-of-the-art methods, including PACE~\citep{ni2024pace}, SoMA~\citep{yun2025soma}, DoRA~\citep{liu2024dora}, LP-FT~\citep{kumar2022finetuning}, AdamSPD~\citep{tian2024rethinking}, CLIPOOD~\citep{shu2023clipood}, CaRot~\citep{oh2024towards}, and FLYP~\citep{goyal2023finetune}. See Appendix~\ref{appx:baselines} for detailed descriptions of all baselines.

Preliminary experiments on DomainNet suggest that, for PEFT methods, updating approximately $10\%$ of parameters provides a favorable adaptation–robustness trade-off. For GMixout, we fix the EMA coefficient to $\lambda = 0.5$ and tune the resampling frequency $k$ based on ID validation performance. A comprehensive analysis of the hyperparameter sensitivity of GMixout is provided in Section~\ref{ablations}. Across all experiments, we replace every linear layer with its corresponding PEFT variant. Additional implementation details are deferred to Appendix~\ref{appx:impl}.

\subsection{Main Results}
\noindent\textbf{Medium-scale Benchmarks.} We begin by evaluating performance on medium-scale benchmarks (40k–500k samples), including DomainNet, iWildCam, CIFAR100-LT and ImageNet-LT (Table~\ref{tab:main_1},~\ref{tab:lt}). Across these datasets, the zero-shot CLIP baseline serves as a proxy for the proximity between the downstream task and the pretraining distribution: DomainNet exhibits the closest alignment, while iWildCam is the most distant.

On DomainNet, GMixout achieves the best overall performance with an average accuracy of 72.6\%. This corresponds to improvements of +5.3 points over Mixout, +2.6 over zero-shot CLIP, +7.6 over Model Soups, and +1.1 over the strongest baseline, while consistently leading across all domains (Real, Sketch, Painting, and Clipart). On iWildCam, where zero-shot performance is substantially lower (ID: 11.5, OOD: 12.7), GMixout yields a large performance gain, achieving the best results both on ID and OOD macro-F1 score and substantially improving over Mixout in both settings. Among baseline methods, CaRot, PACE, and SoMA are the closest competitor, but with important limitations: (i) CaRot requires full-parameter finetuning of both the vision and language encoders, incurring significantly higher computational and memory costs; and (ii) the performance of all three methods strongly depends on the alignment between the pretrained model and the downstream task, which limits their effectiveness on iWildCam. 

On long-tailed benchmarks (ImageNet-LT and CIFAR100-LT; Table~\ref{tab:lt}), GMixout achieves the best performance in both overall and few-shot settings. The gains are particularly pronounced on ImageNet-LT, with improvements of +6.1 and +11.6 over Mixout in the overall and few-shot regimes, respectively, highlighting its robustness on underrepresented classes. Full results across all imbalance categories are reported in Table~\ref{tab:lt-full} in the Appendix.

\begin{table}[ht]
\centering
\caption{OOD accuracy of GMixout and SOTA methods on four DomainNet domains (train on one, evaluate on the others), and macro-F1 on ID and OOD test sets of iWildCam (WILDS).}
\begin{adjustbox}{width=0.7\linewidth}
\begin{tabular}{l|ccccc|cc}
\toprule
\multirow{2}{*}{\textbf{Method}} & \multicolumn{5}{c|}{\textbf{DomainNet (Training Domains)}} & \multicolumn{2}{c}{\textbf{iWildCam}} \\
 & Real & Sketch & Painting & Clipart & Avg. & ID & OOD \\
\midrule
{Zero-shot} & 65.8 & 72.8 & 71.4 & 70.4 & 70.1 & 11.5 & 12.7 \\
\midrule
{Full-FT} & 60.2 & 62.4 & 61.6 & 61.9 & 61.5 & 46.5 & 36.4 \\
{MovingAvg} & 60.2 & 62.3 & 62.0 & 61.8 & 61.6 & 46.6 & 34.2 \\
{Model Soups} & 64.3 & 65.9 & 64.6 & 65.2 & 65.0 & 46.8 & \underline{37.6} \\
\midrule
{Linear Probing} & 62.2 & 64.4 & 60.2 & 62.7 & 62.4 & 31.6 & 24.7 \\
{LoRA} & 66.5 & 71.0 & 69.0 & 69.5 & 69.0 & 49.7 & 35.9 \\
{Random Mask} & 66.8 & 72.0 & 68.9 & 69.8 & 69.4 & 47.9 & 35.2 \\
{Mixout} & 66.5 & 69.8 & 65.0 & 67.7 & 67.3 & 43.5 & 32.9 \\
\midrule
{CaRot} & 67.8 & 73.7 & \underline{72.0} & 71.7 & 71.3 & 43.8 & 31.2 \\
{FLYP} & 57.9 & 63.9 & 65.7 & 67.5 & 63.7 & 42.5 & 30.4 \\
{LP-FT} & 59.3 & 61.2 & 60.3 & 59.3 & 60.0 & 47.4 & 31.8 \\
{AdamSPD} & 66.2 & 68.5 & 66.3 & 67.6 & 67.1 & 47.4 & 36.1 \\
{CLIPOOD} & 60.2 & 62.6 & 61.9 & 61.7 & 61.6 & 41.6 & 24.7 \\
{DoRA} & 66.0 & 70.6 & 68.6 & 69.2 & 68.6 & 50.4 & 35.7 \\
{PACE} & \underline{67.5} & \underline{74.0} & 71.6 & \underline{72.9} & \underline{71.5} & 43.1 & 32.3 \\
{SoMA} & 67.3 & 73.8 & 70.8 & \underline{72.9} & 71.2 & \underline{50.8} & 36.2 \\
\bottomrule
\rowcolor{gray!15}
{\textbf{GMixout}} & \textbf{68.3} & \textbf{75.3} & \textbf{73.5} & \textbf{73.4} & \textbf{72.6} & \textbf{51.8} & \textbf{38.4} \\
\bottomrule
\end{tabular}
\end{adjustbox}
\label{tab:main_1}
\end{table}

\begin{table}[ht]
    \begin{minipage}[t]{.45\linewidth}
    \centering
    \caption{Balanced accuracy of GMixout and SOTA methods on the long-tailed benchmarks ImageNet-LT and CIFAR100-LT.}
    \begin{adjustbox}{width=0.9\linewidth}
    \begin{tabular}{l|cc|cc}
    \toprule
    \multirow{2}{*}{\textbf{Method}} & \multicolumn{2}{c|}{\textbf{ImageNet-LT}} & \multicolumn{2}{c}{\textbf{CIFAR100-LT}} \\
    & All & Few & All & Few \\
    \midrule
    {Zero-shot} & 68.2 & 66.4 & 62.0 & 57.3 \\
    \midrule
    {Full-FT} & 73.1 & 60.4 & 79.6 & 69.3 \\
    {MovingAvg} & 73.0 & 63.1 & 81.0 & 71.9 \\
    {Model Soups} & 76.0 & 65.5 & \underline{82.1} & 73.0 \\
    \midrule
    {Linear Probing} & 74.2 & 67.4 & 70.0 & 60.4 \\
    {LoRA} & 76.7 & 67.4 & 81.1 & \underline{75.7} \\
    {Random Mask} & \underline{76.8} & \underline{70.4} & 81.5 & 74.6 \\
    {Mixout} & 70.9 & 59.3 & 81.5 & 72.9 \\
    \bottomrule
    \rowcolor{gray!15}
    {\textbf{GMixout}} & \textbf{77.0} & \textbf{70.9} & \textbf{82.2} & \textbf{76.0} \\
    \bottomrule
    \end{tabular}
    \end{adjustbox}
    \label{tab:lt}
    \end{minipage}
    \hfill
    \begin{minipage}[t]{.49\linewidth}
    \centering
    \caption{Per-step training overhead of GMixout and baselines with a batch size of 16.}
    \begin{adjustbox}{width=1.0\linewidth}
    \begin{tabular}{l|cccc}
    \toprule
    \multirow{2}{*}{\textbf{Method}} & \#Params $\downarrow$ & FLOPs $\downarrow$ & Latency $\downarrow$ & GPU VRAM $\downarrow$  \\
    & (M) & (T) & (ms) & (G)  \\
    \midrule
    \multicolumn{5}{c}{ViT-B/16} \\
    \midrule
    {Full-FT} & 85.5 & 1.7 & 70.4 & 3.2 \\
    {MovingAvg} & 85.5 & 1.7 & 70.4 & 3.2 \\
    {Model Soups} & 427.5 & 8.5 & 70.4 & 16.0 \\
    {Linear Probing} & 0.5 & 0.6 & 22.5 & 0.5 \\
    {LoRA} & 8.7 & 1.3 & 77.6 & 2.3 \\
    {Random Mask} & 9.0 & 1.1 & 112.0 & 2.5 \\
    \bottomrule
    \rowcolor{gray!15}
    {\textbf{GMixout}} & 9.0 & 1.1 & 123.1 & 2.6 \\
    \midrule
    \multicolumn{5}{c}{ViT-L/14@336} \\
    \midrule
    {LoRA} & 28.3 & 12.7 & 690 & 22.5 \\
    \rowcolor{gray!15}
    \bottomrule
    {\textbf{GMixout}} & 31.0 & 14.2 & 1038 & 23.2 \\
    \bottomrule
    \end{tabular}
    \end{adjustbox}
    \label{tab:bench}
    \end{minipage}
\end{table}

\noindent\textbf{Large-scale Benchmarks.} Table \ref{tab:in_full} reports results on ImageNet-1K (ID) and four natural distribution shifts derived from ImageNet (OOD). As expected, methods that fine-tune on ImageNet-1K substantially improve ID accuracy over zero-shot CLIP: all the baselines reach around 83–84\% ID accuracy, compared to 68.2\% for zero-shot. However, the picture changes under a distribution shift. Zero-shot CLIP remains strong on IN-R (76.4) and IN-A (51.6), while full fine-tuning methods (Full-FT, MovingAvg), including Mixout, degrade to the 40s on IN-A. Model Soups partially restore robustness but require costly multi-model ensembles. PEFT methods offer the best trade-off: slightly lower ID accuracy than Model Soups but consistently higher OOD averages than zero-shot. Among them, GMixout achieves the highest OOD average (60.7), corresponding to relative gains of +0.7 points over Model Soups, +1.3 over LoRA, +0.3 over Random Mask, and +11.6 over Mixout.

Experiments on medium- and large-scale datasets show GMixout is most effective at medium scale, with its advantage persisting, though narrowing, at larger scales. Figure~\ref{fig:in_shots} further reinforces this trend on ImageNet: across all ImageNet subsamples (25–200 shots per class), GMixout maintains a strong ID–OOD balance. Unlike Model Soups, which favors ID accuracy, or MovingAvg, which fluctuates, PEFT methods, especially GMixout, scale smoothly with data. Even at 25 shots, GMixout maintains strong OOD accuracy, and with more data, it matches or surpasses the strongest alternatives.

\begin{minipage}{\textwidth}
\begin{minipage}[b]{0.4\textwidth}
\centering
 \includegraphics[width=1.0\textwidth]{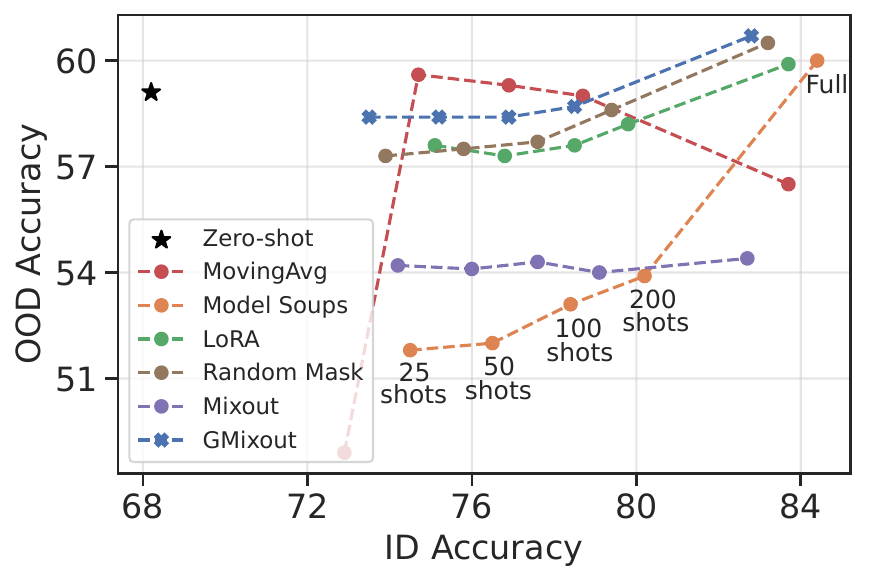}
\captionof{figure}{OOD-ID accuracy on ImageNet with varying training set sizes.}
\label{fig:in_shots}
\end{minipage}
\hfill
\begin{minipage}[b]{0.55\textwidth}
\centering
\begin{adjustbox}{width=1.0\linewidth}
\begin{tabular}{l|c|ccccc}
\toprule
\multirow{2}{*}{\textbf{Method}} & \textbf{ID} & \multicolumn{5}{c}{\textbf{OOD}} \\
            & IN-1k & IN-V2 & IN-R & IN-Sketch & IN-A & Avg. \\
\midrule
{Zero-shot} & 68.2 & 62.0 & \textbf{76.4} & 46.6 & \textbf{51.6} & 59.1 \\
\midrule
{Full-FT} & \underline{83.7} & 73.9 & 65.0 & 47.1 & 40.0 & 56.5 \\
{MovingAvg} & \underline{83.7} & 73.9 & 64.9 & 47.0 & 40.0 & 56.5 \\
{Model Soups} & \textbf{84.4} & \textbf{74.9} & 68.9 & \textbf{50.9} & 45.3 & 60.0 \\
\midrule
{Linear Probing} & 77.0 & 66.4 & 70.0 & 43.7 & 45.7 & 56.5 \\
{LoRA} & \underline{83.7} & \underline{74.0} & 69.3 & 49.6 & 46.6 & 59.9 \\
{Random Mask} & 83.2 & \underline{74.0} & 70.4 & 49.9 & \underline{47.7} & \underline{60.5} \\
{Mixout} & 82.7 & 72.6 & 62.7 & 45.0 & 37.4 & 54.4 \\
\rowcolor{gray!15}
\bottomrule
{\textbf{GMixout}} & 82.8 & 73.6 & \underline{71.6} & \underline{50.4} & 47.2 & \textbf{60.7} \\
\bottomrule
\end{tabular}
\end{adjustbox}
\captionof{table}{Accuracy on ImageNet-1k (ID) and natural ImageNet distribution shifts (OOD).}
\label{tab:in_full}
\end{minipage}
\end{minipage}

\noindent\textbf{Memory-Compute Benchmarks.} Table~\ref{tab:bench} summarizes the per-step training overhead across two architectures (ViT-B/16 and ViT-L/14). Model Soups requires training multiple full models before merging; accordingly, we report its training complexity assuming five parallel runs, corresponding to roughly five dense models. Although GMixout has the same total number of trainable parameters as Full-FT, in our implementation only the unmasked parameters participate in backpropagation, substantially reducing computation and memory per step and making GMixout feasible on limited hardware. LoRA achieves lower latency through better hardware utilization; however, on consumer GPUs, memory remains the primary bottleneck, with usage comparable to GMixout. Finally, all methods share the same inference cost as a single dense baseline, since the learnable parameters in GMixout and LoRA can be merged into $\Phi_0$ after training (see Appendix~\ref{appx:benchmark} for full details).

\subsection{Ablation Studies} \label{ablations}

\begin{minipage}{\textwidth}
\begin{minipage}[b]{0.49\textwidth}
    \centering
    \includegraphics[width=1.0\linewidth]{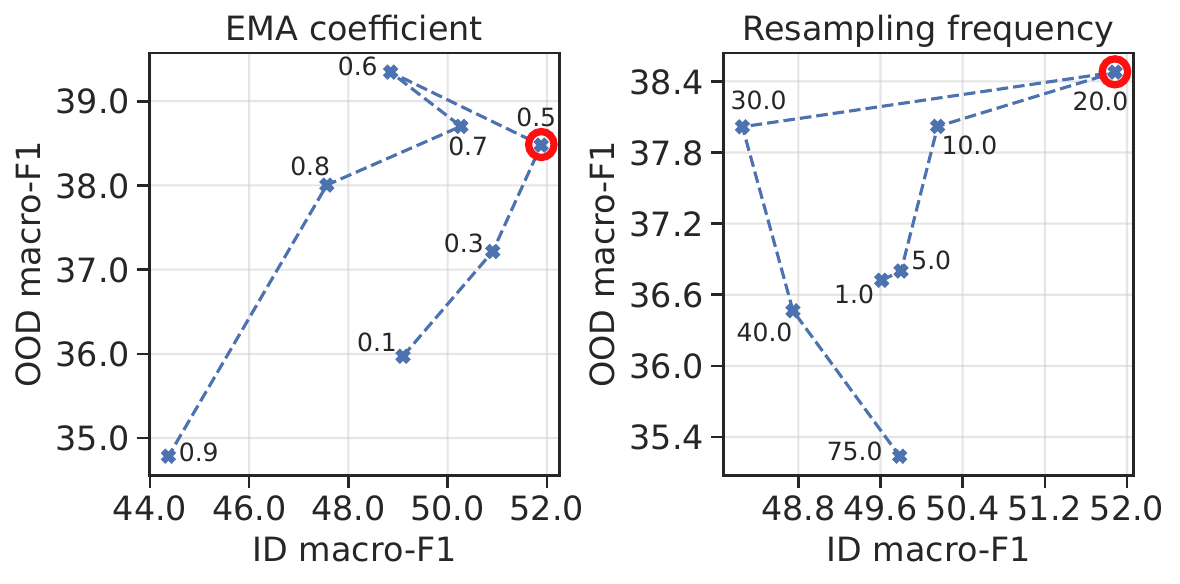}
    \captionof{figure}{OOD-ID curve given variable $\lambda$ (left) and $k$ (right) on the iWildCam dataset.}
    \label{fig:iwild_curve}
\end{minipage}
\hfill
\begin{minipage}[b]{0.49\textwidth}
    \centering
    \begin{adjustbox}{width=0.6\textwidth}
    {
    \begin{tabular}{l|c}
    \toprule
    \textbf{Method} & Worst acc. \\
    \midrule
    Full-FT        & 63.6 \\
    Model Soups    & 62.5 \\
    \midrule
    LoRA           & 58.5 \\
    Random Mask    & 56.8 \\
    Mixout         & 67.6 \\
    DoRA           & 64.5 \\
    \bottomrule
    \rowcolor{gray!15}
    \textbf{GMixout (Ours)} & \textbf{69.0} \\
    \bottomrule
    \end{tabular}
    }
    \end{adjustbox}
    \captionof{table}{Worst-group accuracy on the CivilComments benchmark for language understanding.}
    \label{tab:civil}
\end{minipage}
\end{minipage}

\noindent\textbf{Hyperparameter sensitivity.} 
To further examine how GMixout’s hyperparameters affect the ID--OOD curve, we replicate the analysis from Figure~\ref{fig:intro}(c,d) on iWildCam, a downstream task farther from the pretraining distribution than DomainNet, as reflected by its lower zero-shot OOD performance (12.7 vs.\ 70.1). In Figure~\ref{fig:iwild_curve}, we vary the EMA coefficient $\lambda$ and resampling frequency $k$ separately, keeping mask sparsity fixed at 0.1. Unlike DomainNet, increasing $\lambda$ does not monotonically improve iWildCam performance, and overly large values reduce both ID and OOD accuracy. This supports our analysis in Section~\ref{sec:gmixout_error}: stale anchors can impede adaptation when downstream data are poorly aligned with pretraining. The effect of $k$ is also dataset-dependent, however the values that maximize ID performance also tend to optimize maximize OOD performance, making $k$ easy to tune using ID validation alone. Accordingly, we fix the EMA coefficient to $\lambda = 0.5$ and tune $k$ once for DomainNet and once for iWildCam, and reuse the DomainNet setting for ImageNet, CIFAR, and CivilComments. 

\begin{figure}[ht]
    \centering
    \begin{subfigure}[t]{0.49\linewidth}
     \centering
     \includegraphics[width=1.0\textwidth]{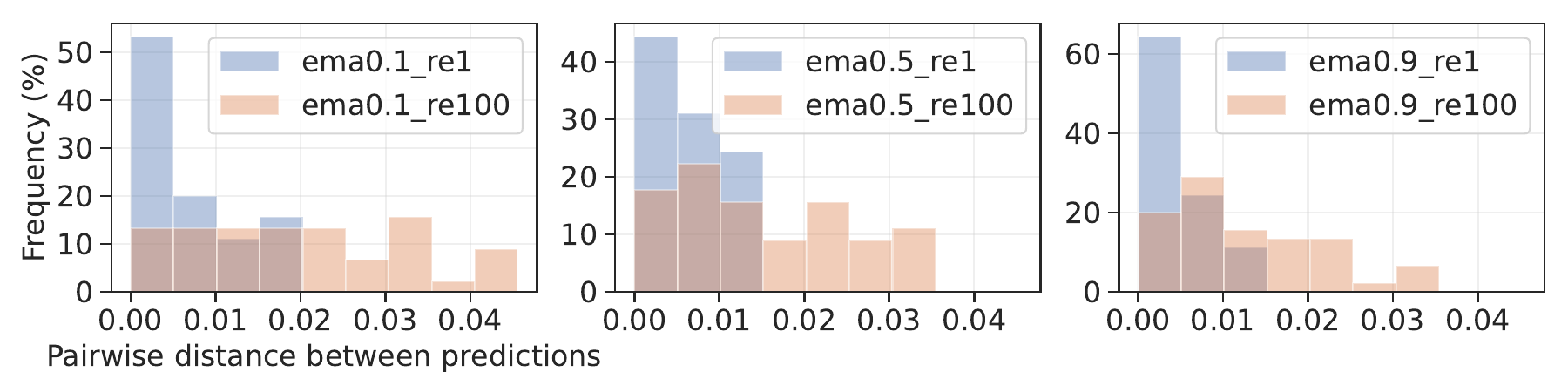}
     \caption{Sketch (ID)}
     \label{fig:pd_dn_sketch}
    \end{subfigure}
    \hfill
    \begin{subfigure}[t]{0.49\linewidth}
        \centering
        \includegraphics[width=1.0\textwidth]{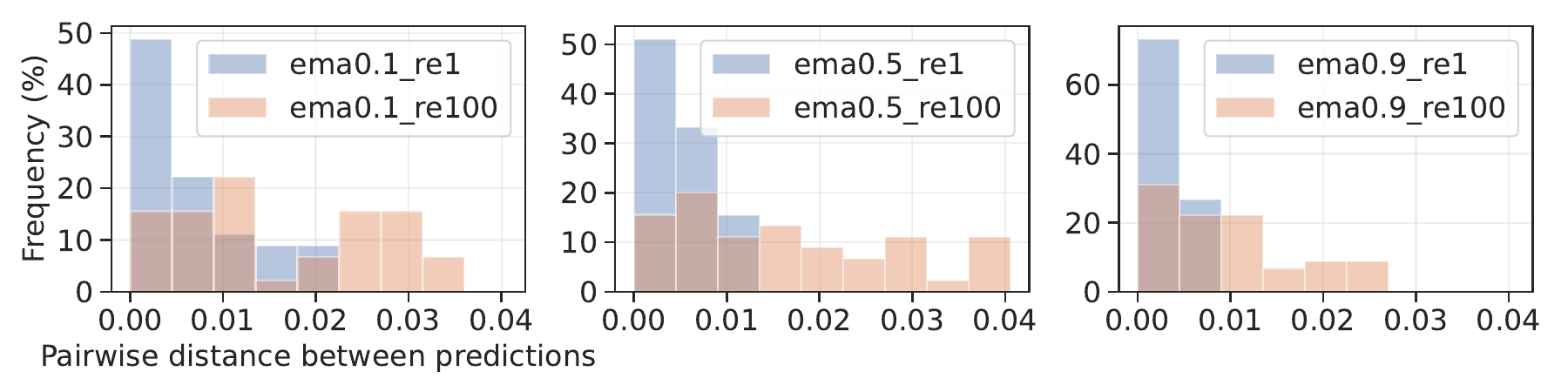}
        \caption{Clipart (OOD)}
        \label{fig:pd_dn_clip}
    \end{subfigure}
    \caption{Frequencies of prediction diversities~\citep{aksela2003comparison} between two subnetworks obtained from a single training run on DomainNet-Sketch and evaluated on Sketch as ID and Clipart as OOD, under $\lambda \in \{ 0.1, 0.5, 0.9 \}$ and $k \in \{1, 100 \}$.}
    \label{fig:pd_dn}
\end{figure}

\noindent\textbf{Prediction diversity.} 
Diversity among ensemble members is a key driver of ensemble effectiveness~\citep{dietterich2000ensemble,rame2022diverse,fort2019deep,izmailov2018averaging}. Figure~\ref{fig:pd_dn} measures pairwise prediction diversity between GMixout checkpoints on DomainNet-Sketch using the ratio-error metric~\citep{aksela2003comparison}. For $k=100$, we extract the weights from the last ten re-masking steps of GMixout. To ensure fair comparison, we extract an equivalent number of checkpoints for the $k=1$ setting over the same training interval. Across ID (Figure~\ref{fig:pd_dn_sketch}) and OOD (Figure~\ref{fig:pd_dn_clip}) evaluations, $k=1$ yields near-zero diversity for many checkpoint pairs, indicating that subnetworks collapse to similar behavior; increasing $k$ substantially increases diversity. Diversity is highest at $\lambda=0.1$ and decreases as $\lambda$ grows, consistent with stronger EMA anchoring suppressing variation across subnetworks (refer to Figure~\ref{fig:pd_dn_full} in the Appendix for results on all domains).

\begin{figure}[ht]
    \centering
    \includegraphics[width=0.8\linewidth]{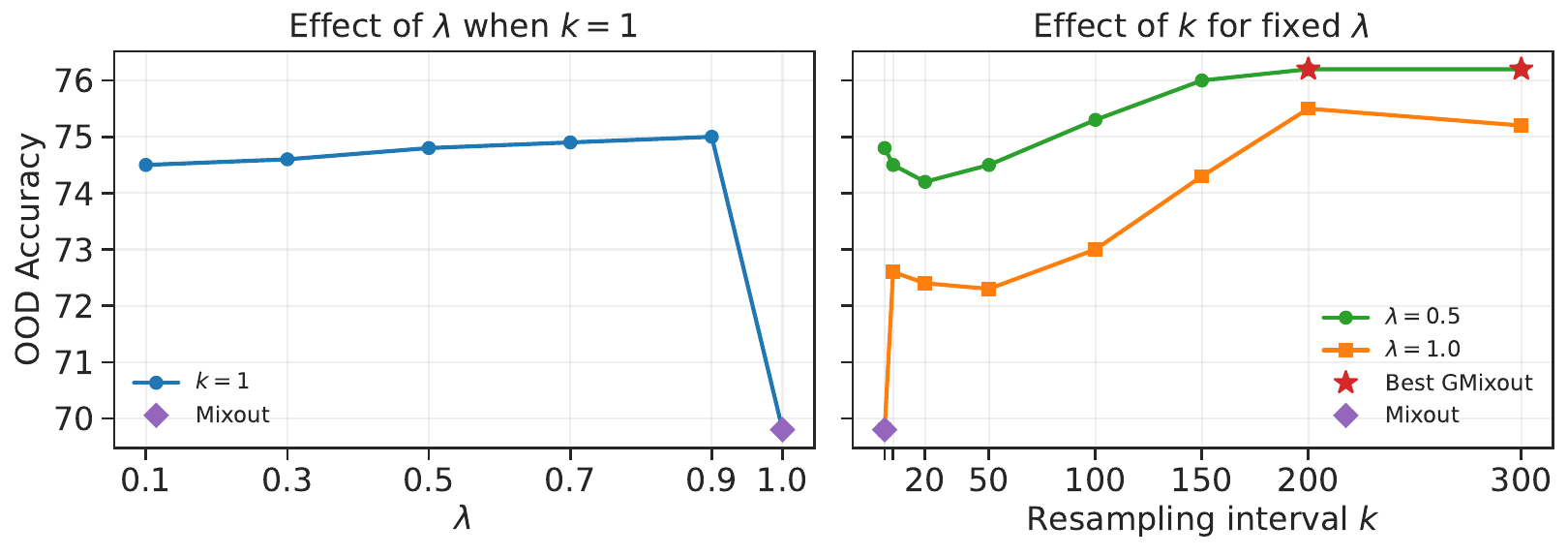}
    \caption{OOD performance of GMixout on DomainNet-Sketch under different choices of $k$ and $\lambda$. The standard Mixout setting corresponds to $k=1$ and $\lambda=1$.}
    \label{fig:appx-isolate}
\end{figure}

\noindent\textbf{Synergy of GMixout components.} To disentangle the contributions of GMixout's components, we conduct ablation experiments on DomainNet-Sketch by varying one hyperparameter at a time while fixing the other to its standard Mixout value, with $k=1$ and $\lambda=1$ corresponding to standard Mixout. Figure~\ref{fig:appx-isolate} presents the results. The left panel isolates the effect of EMA anchoring by varying $\lambda$ while fixing $k=1$, showing that EMA anchoring alone improves performance over standard Mixout by mitigating anchor staleness. Similarly, the right panel isolates the effect of the mask resampling interval by varying $k$ while fixing $\lambda=1$, demonstrating consistent improvements with longer intervals, which reduce correlations among the implicit subnetworks. Combining both modifications yields further gains: the default GMixout configuration, $(k=100,\lambda=0.5)$, achieves 75.3, outperforming either component in isolation. Increasing the resampling interval to $k=200$ or $k=300$, with $\lambda=0.5$, further improves performance to 76.2.

To further distinguish the contribution of masked subnetwork training from EMA anchoring, we introduce a no-mask episodic EMA baseline that follows the same episodic anchor-update schedule as GMixout but keeps all parameters active throughout training. This baseline achieves 68.3, compared with 75.3 for the default GMixout configuration, a difference of 7.0 percentage points. This substantial gap demonstrates that EMA anchoring alone cannot explain GMixout's performance gains and underscores the importance of masked subnetwork training.

Taken together, these results highlight the distinct contributions of masking, EMA anchoring, and extended mask resampling intervals. They further suggest that the improvements from EMA anchoring and longer resampling intervals are complementary, with their combination yielding greater gains than either modification alone.

\begin{table}[ht]
\centering
\caption{OOD accuracy on DomainNet across pretrained weight sources and model scales. Results are reported for models trained on Real/Sketch and evaluated on the remaining domains.}
\begin{adjustbox}{width=0.55\linewidth}
\begin{tabular}{l|cc|cc|cc}
\toprule
\multirow{3}{*}{\textbf{Method}}
& \multicolumn{2}{c|}{\textbf{ViT-B/16}}
& \multicolumn{2}{c|}{\textbf{ViT-B/16}}
& \multicolumn{2}{c}{\textbf{ViT-L/14@336}} \\
& \multicolumn{2}{c|}{\textbf{IN-21k}}
& \multicolumn{2}{c|}{\textbf{DINO}}
& \multicolumn{2}{c}{\textbf{CLIP}} \\
& Real & Sketch
& Real & Sketch
& Real & Sketch \\
\midrule
{Zero-shot}
& - & - & - & - & 74.2 & \underline{79.2} \\
\midrule
{Full-FT}
& 55.1 & 58.0 & - & - & - & - \\
{MovingAvg}
& 55.0 & 59.0 & - & - & - & - \\
{Model Soups}
& \underline{56.4} & 59.4 & - & - & - & - \\
\midrule
{Linear Probing}
& 49.0 & 57.8 & 38.4 & 45.4 & 71.0 & 73.3 \\
{LP-FT}
& - & - & \underline{50.3} & 51.6 & - & - \\
{LoRA}
& 54.9 & \underline{60.0} & 49.8 & \underline{52.0} & \underline{74.7} & 75.7 \\
{Random Mask}
& 54.2 & 59.2 & 49.2 & 51.3 & 73.6 & 76.2 \\
{Mixout}
& 55.8 & 59.4 & 48.6 & 50.9 & - & - \\
\bottomrule
\rowcolor{gray!15}
{\textbf{GMixout}}
& \textbf{65.0} & \textbf{60.3}
& \textbf{52.3} & \textbf{53.6}
& \textbf{76.2} & \textbf{81.4} \\
\bottomrule
\end{tabular}
\end{adjustbox}
\label{tab:ablate_dn}
\end{table}

\noindent\textbf{Beyond standard CLIP backbones.} We further evaluate GMixout by varying the pretrained weight source and model scale. For vision-only ViT-B/16 backbones pretrained on ImageNet-21k~\citep{ridnik2021imagenet} and DINO~\citep{caron2021emerging}, with classification heads initialized from linear probing~\citep{kumar2022finetuning}, GMixout remains competitive despite these models being less robust than CLIP-based ones. As shown in Table~\ref{tab:ablate_dn}, it outperforms Model Soups on ImageNet-21k by 8.6 points on Real and 0.9 points on Sketch, and improves over the best competing DINO baseline by 2.0 and 1.6 points, respectively. To test scalability, we also finetune ViT-L/14@336 CLIP on a single 24GB RTX 3090, where full finetuning is infeasible. GMixout achieves the best performance, surpassing the strongest PEFT baseline by 1.5 points on Real and 5.2 on Sketch, while also improving over zero-shot by 2.2 points on Sketch.

\noindent\textbf{Language understanding.} Although our evaluation focuses primarily on vision benchmarks, GMixout makes no vision-specific assumptions and can be applied to any pretrained model. To illustrate this broader applicability, we include a preliminary language experiment on CivilComments~\citep{duchene2023benchmark}, a standard benchmark for subpopulation shift~\citep{koh2021wilds,yang2023change}. We follow the WILDS data splits~\citep{koh2021wilds} and hide demographic labels during training. Following~\citet{yang2023change}, we finetune BERT$_{\text{BASE}}$~\cite{devlin2018bert} and report worst-group accuracy, a standard measure of robustness to subpopulation imbalance. We apply PEFT layers to all linear layers in the transformer. As shown in Table~\ref{tab:civil}, unlike the vision benchmarks, the standard Mixout is already the strongest finetuning baseline on CivilComments; nevertheless, GMixout further improves worst-group accuracy to 69.0\%, outperforming Mixout by 1.4 points and achieving the best overall robustness.
\section{Conclusions}

We reinterpret Mixout as a single-run implicit ensemble, identifying anchor choice, resampling frequency, and mask sparsity as key drivers of robustness. Building on this, we propose GMixout, which combines an adaptive EMA anchor, controlled resampling, and an efficient sparse implementation. Across five OOD benchmarks, GMixout consistently improves robustness while preserving ID performance, outperforming Mixout, prior PEFT methods, and Model Soups, while also reducing training overhead. Overall, it offers a practical and parameter-efficient approach to robust fine-tuning of vision foundation models.

\paragraph{Acknowledgements.} This work was supported by Distech Controls Inc., the Natural Sciences and Engineering Research Council of Canada, the Digital Research Alliance of Canada, and MITACS.

\bibliography{example_paper}
\bibliographystyle{plainnat}

\newpage
\input{checklist.tex}

\newpage
\appendix

\section{Appendix}

\begin{table*}[ht]
\centering
\caption*{Appendix Contents}
\begin{adjustbox}{width=0.7\textwidth}
\begin{tabular}{ll}
\toprule
\textbf{Section} & \textbf{Page} \\
\midrule
\ref{appx:mixout_wd} Regularization View for Mixout & \pageref{appx:mixout_wd} \\
\ref{appx:full_bvcl} Bias--Variance--Covariance--Locality Decomposition & \pageref{appx:full_bvcl} \\
\ref{appx:forgetting} Measure forgetting & \pageref{appx:forgetting} \\
\ref{appx:param_budget} Parameters budget & \pageref{appx:param_budget} \\
\ref{appx:data_aug} Data Augmentation & \pageref{appx:data_aug} \\
\ref{appx:inet-extend} Extended training on ImageNet-1k & \pageref{appx:inet-extend} \\
\ref{appx:wise-ft} WISE-FT & \pageref{appx:wise-ft} \\
\ref{appx:corruption} Robustness under common corruptions and perturbations & \pageref{appx:corruption} \\
\ref{appx:seeds} Statistical significance analysis & \pageref{appx:seeds} \\
\ref{appx:prompt-tuning} Comparison with Prompt-tuning & \pageref{appx:prompt-tuning} \\
\ref{appx:benchmark} Benchmark details & \pageref{appx:benchmark} \\
\ref{appx:baselines} Additional baselines & \pageref{appx:baselines} \\
\ref{appx:related_work} Extended Related Work & \pageref{appx:related_work} \\
\ref{appx:datasets} Datasets & \pageref{appx:datasets} \\
\ref{appx:impl} Implementation details & \pageref{appx:impl} \\
\ref{sec:impact} Broader impact & \pageref{sec:impact} \\
\ref{appx:full_res} Detailed results & \pageref{appx:full_res} \\
\bottomrule
\end{tabular}
\end{adjustbox}
\end{table*}

\subsection{Regularization View for Mixout}
\label{appx:mixout_wd}

Assume that the loss function $\mathcal{L}$ is strongly convex in a neighborhood of $\Phi_0$ and consider the Mixout parameterization as in Eq.~\ref{eq:mixout}. \citet{lee2019mixout} by a first-order expansion around $\Phi_0$ shows the following inequality:
\begin{equation}
\label{eq:mixout-l2}
\mathbb{E}\bigl[\mathcal{L}\bigl(\Phi_0 + \tfrac{1}{1-p}(M\odot\Delta)\bigr)\bigr]
\;\ge\;
\mathcal{L}(\Phi_0) \;+\; \frac{\mu\, (p)}{2(1-p)}\,\|\Delta\ - \Phi_0||_2^2,
\end{equation}
where $\mu >0$. Eq.~\ref{eq:mixout-l2} shows that, as the mask probability $p$ increases, the effective $\ell_{2}$-penalty grows, biasing the solution toward $\Phi_0$.

\subsection{Bias–variance–covariance–locality Decomposition} \label{appx:full_bvcl}

Consider $M$ finetuned models, each trained under an identically distributed (i.d.) learning procedure $l_{P_{\text{id}}}$. The expected error of a model with weights  $\Phi_{\text{WA}} = \tfrac{1}{M} \sum_{m=1}^M \Phi_m$  with respect to the joint distribution $L_{P_{\text{id}}}^M = \{l_{P_{\text{id}}}^{(m)}\}_{m=1}^M$ was decomposed in~\citet{rame2023model} into:

\begin{align} \label{eq:full_bvcl}
\begin{split}    
\mathbb{E}_{L_{P_{\text{id}}}^M}\bigl[\mc{L}_{\text{ood}}(\Phi_{WA})\bigr]
&=\mathbb{E}_{(x,y)\sim P_{\text{ood}}}\Big[
{\text{bias}^2(x,y)}
+\frac{1}{M}{\text{var}(x)}
+\frac{M-1}{M}{\text{cov}(x)}
\Big]
+ O(\bar{\Pi}^2), \\
where \; \text{bias}(x,y) &= y - \bar{f}(x), \\
and \; \text{var}(x) &= \mathbb{E}_{l_{P_{\text{id}}}^i}\Big[\big(f(x,\Phi_i) - \bar{f}(x)\big)^2\Big], \\
and \; \text{cov}(x) &= \mathbb{E}_{l_{P_{\text{id}}}^i,l_{P_{\text{id}}}^j}\Big[\big(f(x,\Phi_i) - \bar{f}(x)\big)\big(f(x,\Phi_j) - \bar{f}(x)\big)\Big], \\
and \;  \bar{\Pi}^2 &= \mathbb{E}_{L_{P_{\text{id}}}^M}\bigl[\max_{i}\|\Phi_i-\Phi_{WA}\|_2^2\bigr],
\end{split}
\end{align}

where $\bar{f}(x) = \mathbb{E}[f(x,\Phi_i)]$ denotes the expected prediction of $f$, $\text{bias}(x,y)$ and $\text{var}(x)$ are the bias and variance of the models with respect to a sample $(x,y)$, and $\text{cov}(x)$ is the prediction covariance between two ensemble members whose weights are averaged. The locality term $\bar{\Pi}^2$ represents the expected squared maximum distance between individual weights and their weight average.

\subsection{Measure forgetting} \label{appx:forgetting} To study how finetuning alters the backbone’s representations, we evaluate transfer performance on five classification tasks: Flowers102~\citep{Nilsback08}, Food101~\citep{bossard14}, Oxford Pets~\citep{parkhi12a}, DTD~\citep{Cimpoi_2014_CVPR}, and Caltech101~\citep{fei2007learning}. In this setup, the finetuned backbone is frozen and paired with a new text-encoder-based classification head. Figure~\ref{fig:forget} shows that LoRA best preserves prior knowledge, followed by GMixout—an important property, as the goal is to adapt to new tasks without forgetting what has already been learned.

\subsection{Parameters budget} \label{appx:param_budget} Figure~\ref{fig:ratios} compares the effect of the learnable parameter ratio—controlled by rank in LoRA and by mask sparsity in Random Mask, Mixout, and GMixout—on the ID–OOD curve. We evaluate ratios from 1\% to 10\% of learnable parameters. As the ratio increases, all methods trade off OOD for ID, with LoRA being the least sensitive. Across both domains, GMixout consistently yields higher OOD accuracy under any parameter budget while maintaining competitive ID accuracy on Real. On Sketch, it requires lower sparsity to match Random Mask and Mixout. Overall, optimizing about 10\% of parameters is a good heuristic for balancing ID and OOD performance.

\begin{minipage}{\textwidth}
\begin{minipage}[b]{0.4\textwidth}
\centering
 \includegraphics[width=0.7\textwidth]{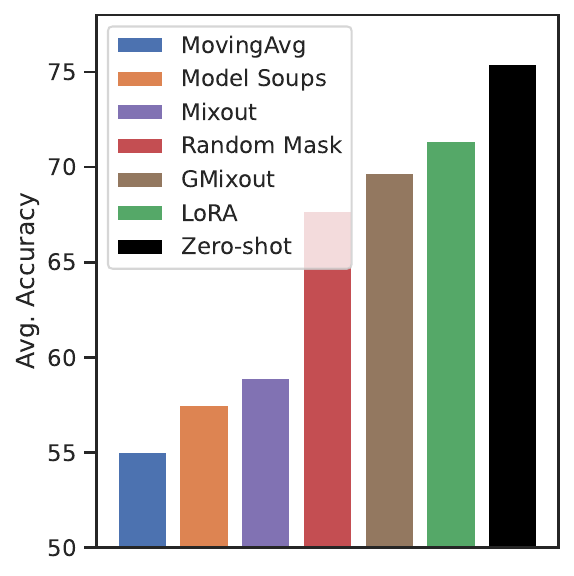}
\captionof{figure}{Average transfer accuracy on five out-of-task datasets after finetuning the methods on ImageNet-1k.}
\label{fig:forget}
\end{minipage}
\hfill
\begin{minipage}[b]{0.55\textwidth}
\centering
\centering
 \includegraphics[width=1.0\textwidth]{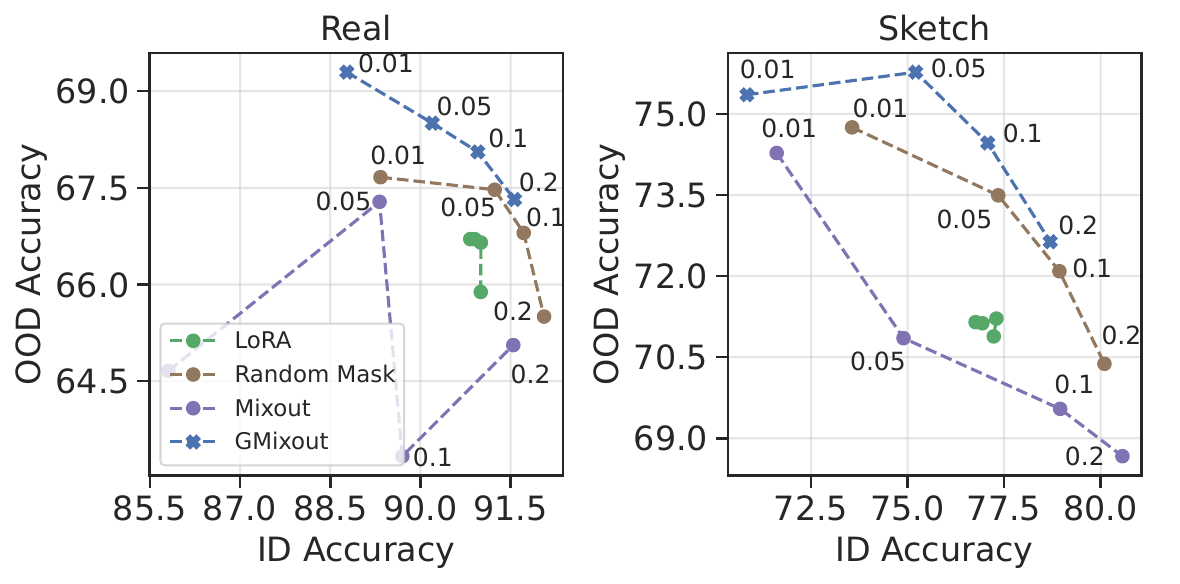}
\captionof{figure}{OOD-ID curve given the variable trainable parameters budget for PEFT methods.}
\label{fig:ratios}
\end{minipage}
\end{minipage}

\subsection{Data Augmentation} \label{appx:data_aug} 
GMixout differs fundamentally from data augmentation as it operates in weight space rather than input space. While techniques like RandAugment~\citep{cubuk2020randaugment} improve robustness by applying random input perturbations—such as cropping, color jittering, and geometric distortions—often relying on domain-specific assumptions about label-preserving transformations, GMixout is instead a domain-agnostic parameter-space regularizer that stochastically reverts subsets of weights to an anchor while controlling subnetwork diversity. As shown in Table~\ref{tab:ra_dn}, data augmentation improves robustness across all baselines, with particularly pronounced gains for GMixout when training on the Sketch domain, yielding a 0.6-point improvement over the standard pipeline—likely due to Sketch’s similarity to other rendition domains such as Painting and Clipart. Overall, augmentation broadens the training distribution, whereas GMixout constrains and diversifies optimization, making the two approaches complementary, consistent with our RandAugment results in Table 8.


\begin{table}[ht]
\centering
\caption{OOD accuracy on DomainNet Real and Sketch with applying \textbf{RandAugment} during finetuning.}
\begin{adjustbox}{width=1.0\linewidth}
\begin{tabular}{l|c|cccc|c|cccc}
\toprule
\multirow{2}{*}{\textbf{Method}} & \textbf{ID} & \multicolumn{4}{c|}{\textbf{OOD}} & \textbf{ID} & \multicolumn{4}{c}{\textbf{OOD}} \\
& Real & Clipart & Sketch & Painting & Avg. & Sketch & Real & Clipart & Painting & Avg. \\
\midrule
{Zero-shot} & 86.1 & 69.3 & 62.0 & 66.1 & 65.8 & 65.4 & 82.9 & 69.3 & 66.1 & 72.8 \\
\midrule
{Full-FT} & 91.7 & 68.0 & 55.0 & 61.2 & 61.4 & \underline{80.7} & 65.9 & 68.3 & 54.5 & 63.0 \\
{MovingAvg} & 91.7 & 67.9 & 55.0 & 61.2 & 61.4 & \underline{80.7} & 66.3 & 68.4 & 54.7 & 63.1 \\
{Model Soups} & 92.2 & 70.8 & 59.6 & 64.7 & 65.0 & \textbf{81.7} & 70.6 & 71.1 & 59.5 & 67.0 \\
\midrule
{Linear Probing} & 90.4 & 66.8 & 59.0 & 63.6 & 63.1 & 74.2 & 72.5 & 64.8 & 57.7 & 65.0 \\
{LoRA} & 90.6 & 72.0 & 63.1 & 66.3 & 67.1 & 76.0 & 79.0 & 71.4 & 65.7 & 72.0 \\
{Random Mask} & 91.5 & 72.3 & 63.1 & 66.3 & 67.3 & 78.0 & 79.6 & 72.5 & 66.0 & \underline{72.7} \\
{Mixout} & 92.0 & 69.7 & 57.9 & 63.3 & 63.6 & 78.9 & 76.5 & 71.9 & 63.3 & 70.6 \\
\bottomrule
\rowcolor{gray!15}
{\textbf{GMixout}} & 90.9 & 72.6 & 63.5 & 66.8 & \textbf{67.7} & 76.2 & 82.4 & 73.5 & 69.3 & \textbf{75.0} \\
\bottomrule
\end{tabular}
\end{adjustbox}
\label{tab:ra_dn}
\end{table}

\begin{table}[ht]
\centering
\caption{OOD accuracy on DomainNet (Real and Sketch) and micro-F1 on iWildCam with \textbf{WISE-FT} with merging coefficient $0.5$.}
\begin{adjustbox}{width=0.6\linewidth}
\begin{tabular}{l|cc|cc|cc}
\toprule
\multirow{3}{*}{\textbf{Method}} & \multicolumn{4}{c|}{\textbf{DomainNet}} & \multicolumn{2}{c}{\textbf{iWildCam}} \\
& \multicolumn{2}{c}{Real} & \multicolumn{2}{c|}{Sketch} & & \\ 
& ID & OOD & ID & OD & ID & OOD \\
\midrule
{Zero-shot} 
& 86.1 & 65.8 & 69.3 & 72.8 & 11.5 & 12.7 \\
\midrule
{Full-FT} 
& \underline{92.0} & 68.9 & 80.4 & 75.2 & 38.6 & 31.1  \\
{MovingAvg} 
& 92.0 & 68.9 & \underline{80.5} & 75.2 & 38.7 & 31.2  \\
{Model Soups} 
& \textbf{92.2} & \underline{69.9} & \textbf{80.7} & 75.7 & 40.1 & \underline{32.3}  \\
\midrule
{LoRA} & 89.0 & 69.0 & 73.1 & \underline{75.6} & 22.4 & 23.2  \\
{Random Mask} & 90.6 & 69.7 & 76.3 & 76.1 & \underline{42.2} & 32.7  \\
{Mixout} & 89.5 & 69.3 & 78.2 & 75.7 & 38.8 & 31.0 \\
\bottomrule
\rowcolor{gray!15}
{\textbf{GMixout}} & 90.1 & \textbf{69.9} & 76.0 & \textbf{76.4} & \textbf{43.1} & \textbf{34.2}  \\
\bottomrule
\end{tabular}
\end{adjustbox}
\label{tab:wise}
\end{table}

\subsection{Extended training on ImageNet-1k} \label{appx:inet-extend} An easy way to improve downstream performance is to train for longer. However, this often reduces robustness, particularly when starting from a strong foundation model. In Table~\ref{tab:in_epochs}, we extend training on ImageNet-1k for 30 epochs. As shown, LoRA and Random Mask show no ID improvement, suggesting overfitting, while GMixout continues to improve ID accuracy. OOD accuracy drops across all methods, but GMixout is the most resilient. A similar trend holds for out-of-task datasets, where all methods experience decreased performance.

\begin{table}[ht]
\centering
\caption{Accuracy on ImageNet-1k (ID) and four natural distribution shifts (OOD) under extended finetuning, along with transfer performance on five out-of-task datasets.}
\begin{adjustbox}{width=0.8\linewidth}
\begin{tabular}{l|c|ccccc|c}
\toprule
\multirow{2}{*}{\textbf{Method}} & \textbf{ID} & \multicolumn{5}{c|}{\textbf{OOD}} & \textbf{Out-of-Task} \\
            & IN-1k & IN-V2 & IN-R & IN-Sketch & IN-A & Avg. & Avg. of 5 Datasets \\
\midrule
{Zero-shot} & 68.2 & 62.0 & 76.4 & 46.6 & 51.6 & 59.1 & 75.4\\
\midrule
\multicolumn{8}{c}{\textbf{10 Epochs}} \\
\midrule
{LoRA} & 83.7 & 74.0 & 69.3 & 49.6 & 46.6 & 59.9 & 71.4 \\
{Random Mask} & 83.2 & 74.0 & 70.4 & 49.9 & 47.7 & 60.5 & 67.7 \\
{Mixout} & 82.7 & 72.6 & 62.7 & 45.0 & 37.4 & 54.4 & 58.9 \\
\rowcolor{gray!15}
\bottomrule
{\textbf{GMixout}} & 82.8 & 73.6 & 71.6 & 50.4 & 47.2 & 60.7 & 69.7 \\
\midrule
\multicolumn{8}{c}{\textbf{30 Epochs}} \\
\midrule
{LoRA} & 83.5 & 74.0 & 68.9 & 49.6 & 47.2 & 59.9 & 70.2 \\
{Random Mask} & 83.2 & 73.8 & 69.1 & 49.2 & 47.3 & 59.9 & 66.4 \\
{Mixout} & 81.8 & 71.2 & 58.4 & 41.3 & 32.4 & 50.8 & 55.5 \\
\bottomrule
\rowcolor{gray!15}
{\textbf{GMixout}} & 83.1 & 73.8 & 70.4 & 50.1 & 46.1 & 60.1 & 68.0 \\
\bottomrule
\end{tabular}
\end{adjustbox}
\label{tab:in_epochs}
\end{table}

\subsection{WISE-FT} \label{appx:wise-ft} An effective way to combine the downstream performance of finetuned models with the robustness of zero-shot models is to ensemble their weights~\citep{wortsman2022robust}. This technique, known as WISE-FT, can be applied to any finetuning method. By adjusting a merging coefficient, WISE-FT trades downstream adaptability for robustness. Table~\ref{tab:wise} reports results on DomainNet (trained on Real/Sketch) and iWildCam with a coefficient of $0.5$. On DomainNet, where the zero-shot model is already strong, WISE-FT improves all baselines. However, on iWildCam, where CLIP performs poorly in zero-shot, WISE-FT provides no benefit—in fact, the baselines perform better without it. In contrast, GMixout (without WISE-FT) consistently improves OOD performance regardless of the zero-shot model’s strength.

\subsection{Robustness under common corruptions and perturbations} \label{appx:corruption}
We further compare GMixout with baselines under synthetic corruptions using \texttt{CIFAR100-C}~\citep{hendrycks2019benchmarking}, which introduces 15 corruption types at the highest severity level (5). Table~\ref{tab:cifar100} summarizes the results.

On CIFAR100, where CLIP’s transfer to low-resolution images is more limited, GMixout achieves the second-best performance under both clean and corrupted settings, trailing Model Soups by only 0.2 and 0.9 points, respectively, while outperforming all other methods.

\begin{table}[ht]
\centering
\caption{Accuracy on CIFAR100 (ID) and CIFAR100-C (Corruption).}
\begin{adjustbox}{width=0.4\linewidth}
\begin{tabular}{l|cc}
\toprule
\textbf{Method} & \textbf{ID} & \textbf{Corruption} \\
\midrule
{Zero-shot} & 64.3 & 33.3 \\
\midrule
{Full-FT} & 89.7 & 57.6  \\
{MovingAvg} & 89.9 & 58.7  \\
{Model Soups} & \textbf{91.1} & \textbf{63.0}  \\
\midrule
{Linear Probing} & 74.1 & 42.9  \\
{LoRA} & 89.9 & 55.0  \\
{Random Mask} & 90.2 & 56.1  \\
{Mixout} & 88.7 & 55.7  \\
\midrule
{LP-FT} & 74.6 & 43.0  \\
{AdamSPD} & 88.2 & 56.5  \\
{CLIPOOD} & 86.7 & 53.2  \\
{DoRA} & 88.2 & 54.4  \\
{PACE} & 87.4 & 54.7 \\
{SoMA} & 87.8 & 55.2 \\
\midrule
\rowcolor{gray!15}
{\textbf{GMixout}} & \underline{90.9} & \underline{62.1}  \\
\bottomrule
\end{tabular}
\end{adjustbox}
\label{tab:cifar100}
\end{table}

\subsection{Statistical significance analysis} \label{appx:seeds}

\begin{figure}[ht]
    \centering
    \includegraphics[width=0.8\linewidth]{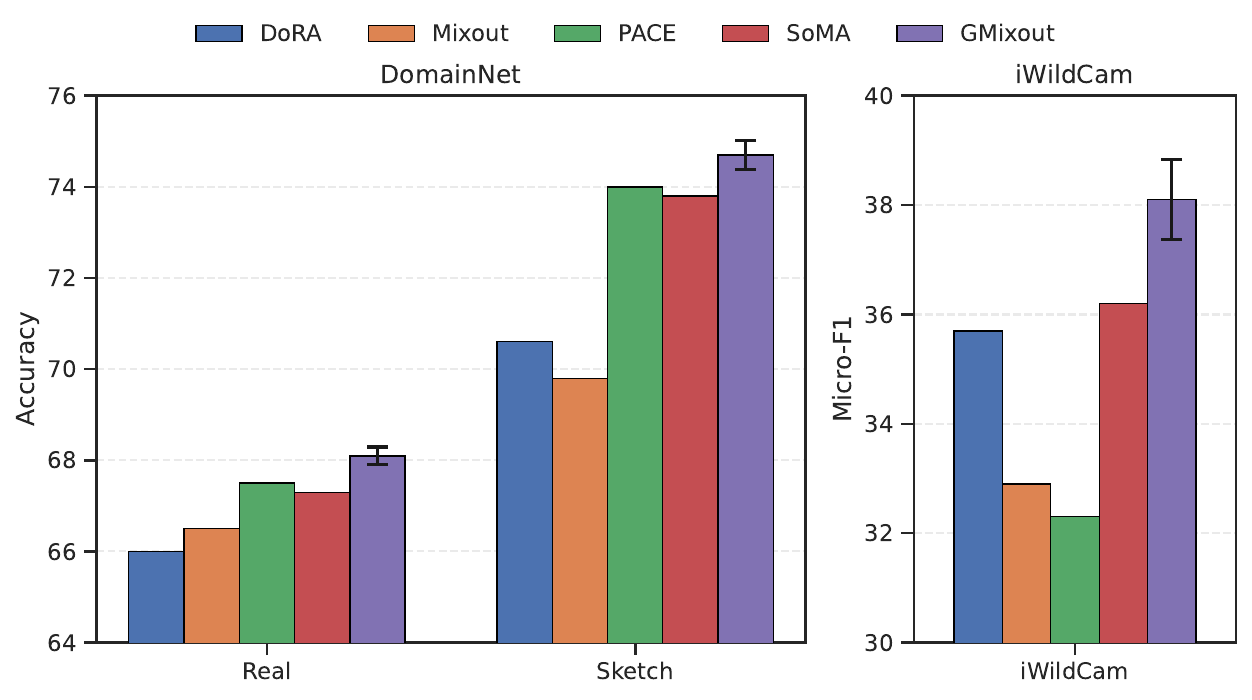}
    \caption{OOD accuracy on DomainNet (trained on Real/Sketch) and OOD micro-F1 on iWildCam, across 3 seeds.}
    \label{fig:seed}
\end{figure}

We evaluate GMixout across three random seeds on three benchmarks: DomainNet-Real, DomainNet-Sketch, and iWildCam. As shown in Figure~\ref{fig:seed}, GMixout consistently outperforms the compared baselines while maintaining small variance across seeds. This suggests that the gains are not due to a particular randomness but reflect stable improvements across datasets.

\subsection{Comparison with Prompt-tuning} \label{appx:prompt-tuning}
In this section, we compare GMixout with prompt-tuning baselines, including CoOp~\citep{zhou2022learning} and CoCoOp~\citep{zhou2022conditional}, on the few-shot learning benchmark. CoOp replaces hand-crafted prompts in vision–language models, such as CLIP, with learnable context vectors trained end-to-end, thereby improving performance but showing limited generalization to unseen classes. CoCoOp extends this by employing a meta-network to generate image-conditioned context tokens, enhancing generalization at the cost of slower inference. Following the evaluation protocol of~\citet{zhou2022conditional}, we sample 16 shots per class from ImageNet-1k for training and evaluate on both the ID validation set and four natural distribution shifts derived from ImageNet-1k. We use ViT-B/16 with CLIP weights as the image encoder. We use the mask sparsity of $0.01$ for GMixout. As shown in Table~\ref{tab:prompt-tuning}, GMixout consistently outperforms prompt-tuning baselines on both ID and OOD, except on IN-A.

\begin{table}[ht]
\centering
\caption{Accuracy on few-shot ImageNet-1k (ID, 16 shots per class) and four natural distribution shifts (OOD).}

\begin{adjustbox}{width=0.7\linewidth}
\begin{tabular}{l|c|ccccc}
\toprule
\multirow{2}{*}{\textbf{Method}} & \textbf{ID} & \multicolumn{5}{c}{\textbf{OOD}} \\
            & IN-1k & IN-V2 & IN-R & IN-Sketch & IN-A & Avg. \\
\midrule
{Zero-shot} & 68.2 & 62.0 & 76.4 & 46.6 & 51.6 & 59.1 \\
\midrule
{CoOp~\citep{zhou2022learning}} & 71.5 & 64.2 & 75.2 & 48.0 & 49.7 & 59.3 \\
{CoCoOp~\citep{zhou2022conditional}} & 71.0 & 64.0 & 76.2 & 48.7 & 50.6 & 59.9 \\
\rowcolor{gray!15}
\bottomrule 
{\textbf{GMixout}} & \textbf{73.4} & \textbf{65.8} & \textbf{77.3} & \textbf{49.0} & 49.3 & \textbf{60.3} \\
\bottomrule
\end{tabular}
\end{adjustbox}
\label{tab:prompt-tuning}
\end{table}

\subsection{Benchmark details} \label{appx:benchmark}
All benchmarks are conducted using the PyTorch v2.7.1 profiling tools~\citep{paszke2019pytorch}. Experiments are run on an NVIDIA RTX 3090 (24 GB VRAM). We set the LoRA rank to $r=64$, and use a mask sparsity of $s=0.1$ for Random Masking, and GMixout. For ViT-L/14, full finetuning is no longer feasible on a single consumer GPU, so we compare only against LoRA.

The standard Mixout method does not reduce memory usage, since it zeroes gradients only during the backward pass while still requiring all gradients to be stored. As a result, its per-step cost is comparable to full fine-tuning. However, our sparse-kernel implementation is compatible with Mixout and can achieve similar computational benefits to GMixout.


\subsection{Additional baselines} \label{appx:baselines}
In this work, we include a unified set of baselines that are widely recognized for achieving state-of-the-art performance under different types of distribution shift. In particular, for corruption and covariate-shift benchmarks—especially when the target distribution departs substantially from the pretraining data or when datasets are large—Model Soups~\citep{wortsman2022model} constitute a strong and competitive reference~\citep{rame2022diverse,fort2019deep}. Several recent works~\citep{jang2025model,rame2023model} explicitly acknowledge Model Soups as a robustness upper bound and aim to approximate their performance while reducing the associated computational cost. Below, we provide additional details on the state-of-the-art baselines included for the DomainNet, iWildCam, and CIFAR100 benchmarks.

\begin{itemize}
    \item \textbf{DoRA}~\citep{liu2024dora} is a variant of LoRA that decomposes pretrained weights into magnitude and direction, applying low-rank updates primarily to the directional component to better approximate full finetuning capacity without inference-time overhead.
    
    \item \textbf{LP-FT}~\citep{kumar2022finetuning} is a two-stage finetuning strategy that first performs linear probing and then finetunes the model, motivated by the observation that direct finetuning can distort pretrained representations and harm OOD generalization.
    
    \item \textbf{AdamSPD}~\citep{tian2024rethinking} improves robustness by introducing selective projection decay into Adam-style optimizers, strongly regularizing sensitive parameter subspaces while allowing others to adapt during finetuning.
    
    \item \textbf{CLIPOOD}~\citep{shu2023clipood} is a CLIP-specific robust finetuning method that preserves OOD performance by combining margin-based objectives with temporal ensembling to stabilize adaptation under distribution shift.
    
    \item \textbf{CaRot}~\citep{oh2024towards} is a vision--language finetuning approach that jointly targets OOD robustness and confidence calibration, improving both predictive accuracy and calibration under domain shift.
    
    \item \textbf{FLYP}~\citep{goyal2023finetune} (\emph{Finetune Like You Pretrain}) adapts vision--language models by continuing contrastive learning with class prompts, aligning finetuning with the original pretraining objective to improve robustness.
\end{itemize}

\subsection{Extended Related Work} \label{appx:related_work}

Domain generalization (DG) aims to train models on limited source domains that generalize to unseen targets~\citep{zhou2022domain}. Existing approaches can be broadly categorized into: (i) feature alignment methods that learn domain-invariant representations~\citep{ganin2016domain,li2018domain,sun2016deep}, (ii) invariant predictor learning that enforces consistency of classifiers across domains~\citep{arjovsky2019invariant,shi2021gradient}, (iii) distributionally robust optimization that focuses on worst-case performance~\citep{sagawa2019distributionally}, (iv) meta-learning strategies for rapid adaptation to domain shifts~\citep{bui2021exploiting}, and (v) data augmentation methods that increase training diversity~\citep{aminbeidokhti2024domain,cubuk2020randaugment,hendrycks2019augmix,zhang2017mixup}. Despite strong theoretical motivation, many DG methods do not consistently outperform empirical risk minimization (ERM) under fair model selection~\citep{gulrajani2020search,koh2021wilds}, whereas ensemble-based approaches often yield substantial improvements~\citep{wortsman2022model,rame2023model}.

Although related in spirit, DG differs from robust fine-tuning due to its underlying assumptions. DG trains on multiple labeled source domains and evaluates on a held-out target domain. In contrast, robust fine-tuning assumes access to a single or mixed (unlabeled) source domain and evaluates robustness to distribution shifts within the same task. While DG was highly effective in the ImageNet pretraining era, robust fine-tuning is more suitable for evaluating modern foundation models, which already exhibit strong zero-shot generalization on DomainNet-like benchmarks. Notably, recent follow-up work to DomainBed~\citep{gulrajani2020search} adopts a similar protocol for evaluating such models~\citep{mayilvahanan2025in}.

\subsection{Datasets} \label{appx:datasets}

We evaluate on \texttt{ImageNet-1K}~\citep{russakovsky2015imagenet}, with OOD tests on \texttt{ImageNet-V2}~\citep{recht2019imagenet}, \texttt{ImageNet-R}~\citep{hendrycks2021many}, \texttt{ImageNet-Sketch}~\citep{wang2019learning}, and \texttt{ImageNet-A}~\citep{hendrycks2021natural}. From the WILDS benchmark~\citep{koh2021wilds}, we include \texttt{iWildCam}~\citep{beery2021iwildcam}, which contains animal images with variations in camera types, backgrounds, and illumination. \texttt{DomainNet}~\citep{peng2019moment} is also considered, where one domain is treated as ID and the remaining three as OOD. For long-tailed class imbalance, we evaluate on \texttt{CIFAR100-LT}~\citep{cao2019learning} and \texttt{ImageNet-LT}~\citep{liu2019large}. In both datasets, the number of training samples per class follows a long-tailed distribution that decreases exponentially, while the test set is balanced with a uniform distribution across classes. 

\subsection{Implementation details} \label{appx:impl}
As the feature extractor, we use CLIP's vision encoder~\citep{radford2021learning} implemented with a ViT~\citep{dosovitskiy2020image}. For the classification head $g$, a prototypical head is adopted with $\ell_2$-normalized features and class prototypes, and a temperature scaling on the logits. The parameters $w$ are initialized via the CLIP text encoder. The full set of parameters is denoted as $\Phi = \{\theta, w\}$. 

We train all methods with the AdamW optimizer~\citep{loshchilov2018decoupled}, using weight decay $0.1$. Batch size is $128$ for all datasets except ImageNet-1k, where it is $512$. To initialize the linear classifier in CLIP-based ImageNet-1k experiments, we use the prompt templates provided in~\citet{radford2021learning}. For all other datasets, we adopt the template ``a photo of a \texttt{<class>}'' to obtain the corresponding embeddings from the CLIP text encoder. For each baseline, we tune learning rates from $\{ 1e-2, 1e-3, 1e-4, 1-5 \}$ with a cosine decay scheduler that anneals to $0$ after a 1-epoch warmup. Training runs for $10$ epochs on ImageNet-1k and DomainNet, and $20$ epochs on all other datasets. The best checkpoint is selected using each dataset’s in-domain validation set~\citep{gulrajani2020search}.

For long-tail benchmarks, we adopt the \emph{Logit Adjustment} (LA) loss~\citep{menon2020long}, which mitigates head-class bias by adding a label-dependent offset to the logits:
\begin{equation} \label{la_loss}
    \ell_{LA}(y,f(\boldsymbol x)) = - \log \frac{\exp({g_{y}(\boldsymbol x) + \log \pi_y)}}{\sum_{y' \in \mathcal{C}}  \exp({f_{y'}(\boldsymbol x) + \log \pi_{y'})}},
\end{equation}
where $\pi \in \Delta_y$ are estimates of the class priors $P(y)$ based on the empirical class frequencies on the training data $D$. In a recent study, \citet{shi2024long} observed that starting from CLIP pretrained weights, the LA loss alone is insufficient to achieve strong performance. 

\subsection{Broader impact} \label{sec:impact}
This work advances research on robust finetuning under distribution shift, a challenge that commonly arises in real-world applications such as medical imaging, wildlife monitoring, and autonomous driving. By improving robustness without increasing inference cost or training complexity, GMixout supports the deployment of more reliable models in settings where data distributions evolve over time. Improved robustness can also help reduce performance disparities across underrepresented groups or conditions, contributing to fairer and more dependable AI systems.

\subsection{Detail results} \label{appx:full_res}

In this section, we show detailed results of our main experiments. Table~\ref{tab:lt-full} reports the full results on long-tail benchmarks, Tables~\ref{tab:dn_real_sketch} and ~\ref{tab:dn_painting_clipart} report the full results on the Real, Sketch, Clipart, and Painting domains of DomainNet, treating each as ID with their corresponding OOD domains. Table~\ref{tab:in_all_shots} provides the detailed ID–OOD results corresponding to Figure~\ref{fig:in_shots}, using the ImageNet-1k subsampled training sets with 25, 50, 100, and 200 images per class, respectively.

\begin{figure}
    \centering
    \begin{subfigure}[t]{0.49\linewidth}
     \centering
     \includegraphics[width=1.0\textwidth]{figures/pd_domainnet_sketch.pdf}
     \caption{Sketch (ID)}
    \end{subfigure}
    \hfill
    \begin{subfigure}[t]{0.49\linewidth}
        \centering
        \includegraphics[width=1.0\textwidth]{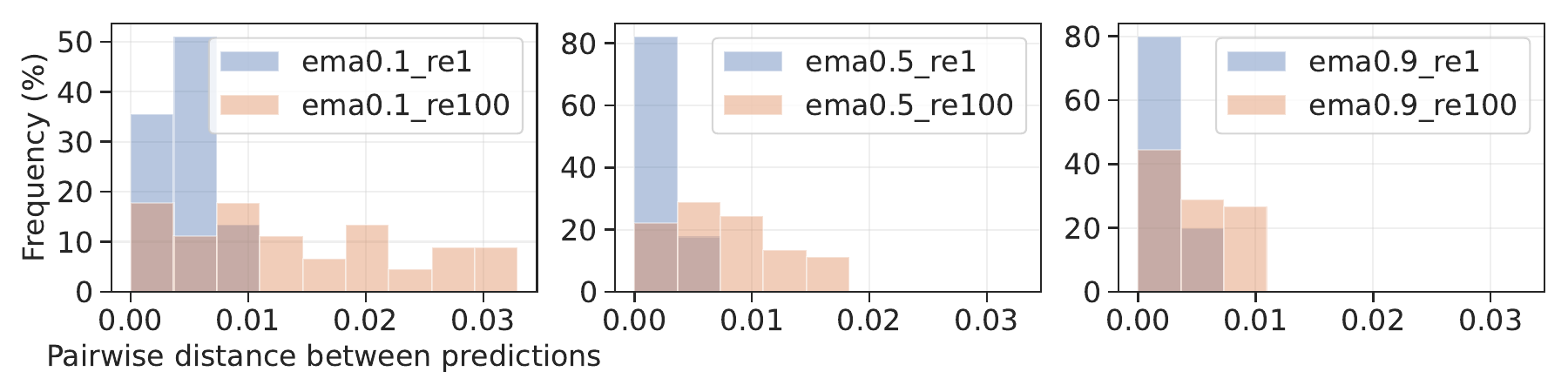}
        \caption{Real (OOD)}
    \end{subfigure}
    \begin{subfigure}[t]{0.49\linewidth}
        \centering
        \includegraphics[width=1.0\textwidth]{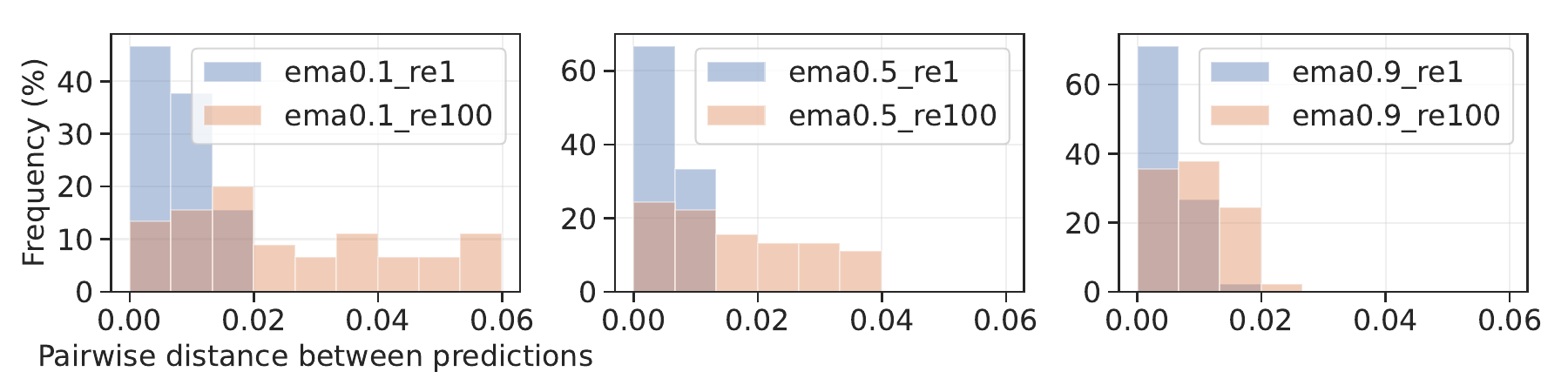}
        \caption{Paint (OOD)}
    \end{subfigure}
    \begin{subfigure}[t]{0.49\linewidth}
        \centering
        \includegraphics[width=1.0\textwidth]{figures/pd_domainnet_clipart.pdf}
        \caption{Clipart (OOD)}
    \end{subfigure}
    \caption{Frequencies of prediction diversities~\citep{aksela2003comparison} between two subnetworks obtained from a single training run on DomainNet-Sketch and evaluated on the rest of domains, under $\lambda \in \{ 0.1, 0.5, 0.9 \}$ and $k \in \{1, 100 \}$.}
    \label{fig:pd_dn_full}
\end{figure}

\begin{table}
\centering
\caption{Balanced accuracy of GMixout and SOTA methods on the long-tailed benchmarks ImageNet-LT and CIFAR100-LT.}
\begin{adjustbox}{width=0.7\linewidth}
\begin{tabular}{l|cccc|cccc}
\toprule
\multirow{2}{*}{\textbf{Method}} & \multicolumn{4}{c|}{\textbf{ImageNet-LT}} & \multicolumn{4}{c}{\textbf{CIFAR100-LT}} \\
& All & Many & Med. & Few & All & Many & Med. & Few \\
\midrule
{Zero-shot} & 68.2 & 69.4 & 67.8 & 66.4 & 62.0 & 66.2 & 61.8 & 57.3 \\
\midrule
{Full-FT} & 73.1 & 80.3 & 70.8 & 60.4 & 79.6 & 88.1 & 79.9 & 69.3 \\
{MovingAvg} & 73.0 & 79.0 & 71.0 & 63.1 & 81.0 & \underline{89.4} & 80.6 & 71.9 \\
{Model Soups} & 76.0 & \underline{81.5} & 74.5 & 65.5 & \underline{82.1}  & \textbf{89.9} & \underline{82.2} & 73.0 \\
\midrule
{Linear Probing} & 74.2 & 77.8 & 73.3 & 67.4 & 70.0 & 77.2 & 71.1 & 60.4 \\
{LoRA} & 76.7 & \textbf{81.6} & 75.3 & 67.4 & 81.1 & 85.7 & 81.0 & \underline{75.7} \\
{Random Mask} & \underline{76.8} & 80.7 & \underline{75.5} & 70.4 & 81.5 & 87.4 & 81.7 & 74.6 \\
{Mixout} & 70.9 & 76.5 & 69.7 & 59.3 & 81.5 & 88.1 & 82.3 & 72.9 \\
\bottomrule
\rowcolor{gray!15}
{\textbf{GMixout}} & \textbf{77.0} & 80.3 & \textbf{76.2} & \textbf{70.9} & \textbf{82.2} & 87.3 & \textbf{82.5} & \textbf{76.0} \\
\bottomrule
\end{tabular}
\end{adjustbox}
\label{tab:lt-full}
\end{table}

\begin{table}
\centering
\caption{OOD accuracy on DomainNet Real and Sketch.}
\begin{adjustbox}{width=1.0\linewidth}
\begin{tabular}{l|c|cccc|c|cccc}
\toprule
\multirow{2}{*}{\textbf{Method}} & \textbf{ID} & \multicolumn{4}{c|}{\textbf{OOD}} & \textbf{ID} & \multicolumn{4}{c}{\textbf{OOD}} \\
& Real & Clipart & Sketch & Painting & Avg. & Sketch & Real & Clipart & Painting & Avg. \\
\midrule
{Zero-shot} & 86.1 & 69.3 & 62.0 & 66.1 & 65.8 & 65.4 & 82.9 & 69.3 & 66.1 & 72.8 \\
\midrule
{Full-FT} & 91.6 & 67.0 & 53.3 & 60.2 & 60.2 & 80.7 & 65.7 & 67.7 & 53.7 & 62.4 \\
{MovingAvg} & 91.6 & 67.0 & 53.4 & 60.2 & 60.2 & 80.7 & 65.7 & 67.8 & 53.5 & 62.3 \\
{Model Soups} & 92.4 & 70.5 & 58.2 & 64.1 & 64.3 & 81.9 & 69.0 & 70.7 & 58.0 & 65.9 \\
\midrule
{Linear Probing} & 90.7 & 66.0 & 57.9 & 62.7 & 62.2 & 75.7 & 71.4 & 65.4 & 56.4 & 64.4 \\
{LoRA} & 90.9 & 71.7 & 62.3 & 65.6 & 66.5 & 76.9 & 77.8 & 71.5 & 63.6 & 71.0 \\
{Random Mask} & 91.7 & 72.1 & 62.2 & 66.1 & 66.8 & 78.9 & 78.8 & 72.9 & 64.6 & 72.0 \\
{Mixout} & 89.7 & 71.5 & 62.4 & 65.7 & 66.5 & 79.1 & 75.2 & 71.3 & 62.8 & 69.8 \\
\bottomrule
\rowcolor{gray!15}
{\textbf{GMixout}} & 91.0 & 73.2 & 64.1 & 67.6 & \textbf{68.3} & 77.0 & 82.6 & 74.1 & 69.3 & \textbf{75.3} \\
\bottomrule
\end{tabular}
\end{adjustbox}
\label{tab:dn_real_sketch}
\end{table}

\begin{table}
\centering
\caption{OOD accuracy on DomainNet Painting and Clipart.}
\begin{adjustbox}{width=1.0\linewidth}
\begin{tabular}{l|c|cccc|c|cccc}
\toprule
\multirow{2}{*}{\textbf{Method}} & \textbf{ID} & \multicolumn{4}{c|}{\textbf{OOD}} & \textbf{ID} & \multicolumn{4}{c}{\textbf{OOD}} \\
& Painting & Clipart & Sketch & Real & Avg. & Clipart & Sketch & Painting & Real & Avg. \\
\midrule
{Zero-shot} & 69.3 & 69.3 & 62.0 & 82.9 & 71.4 & 69.5 & 62.0 & 66.1 & 82.9 & 70.4 \\
\midrule
{Full-FT} & 83.6 & 62.1 & 51.2 & 71.6 & 61.6 & 85.0 & 57.9 & 55.0 & 72.7 & 61.9 \\
{MovingAvg} & 83.6 & 63.0 & 51.5 & 71.7 & 62.0 & 84.9 & 58.0 & 54.9 & 72.5 & 61.8 \\
{Model Soups} & 84.7 & 65.2 & 54.9 & 73.6 & 64.6 & 86.3 & 61.6 & 58.7 & 75.1 & 65.2 \\
\midrule
{Linear Probing} & 80.6 & 58.4 & 53.2 & 68.9 & 60.2 & 81.2 & 57.8 & 56.8 & 73.6 & 62.7 \\
{LoRA} & 81.5 & 68.1 & 61.4 & 77.6 & 69.0 & 81.8 & 64.0 & 64.1 & 80.4 & 69.5 \\
{Random Mask} & 82.8 & 68.4 & 59.9 & 78.3 & 68.9 & 83.2 & 64.5 & 64.5 & 80.4 & 69.8 \\
{Mixout} & 82.5 & 65.1 & 55.8 & 74.3 & 65.0 & 84.9 & 62.3 & 60.7 & 77.6 & 66.9 \\
\bottomrule
\rowcolor{gray!15}
{\textbf{GMixout}} & 81.3 & 71.9 & 66.0 & 82.8 & \textbf{73.5} & 81.8 & 67.6 & 69.0 & 83.7 & \textbf{73.4} \\
\bottomrule
\end{tabular}
\end{adjustbox}
\label{tab:dn_painting_clipart}
\end{table}

\begin{table}
\centering
\caption{ID and OOD results on ImageNet-1k subsampled training datasets (25, 50, 100, and 200)-shots.}
\begin{adjustbox}{width=0.7\linewidth}
\begin{tabular}{l|c|ccccc}
\toprule
\multirow{2}{*}{\textbf{Method}} & \textbf{ID} & \multicolumn{5}{c}{\textbf{OOD}} \\
                        & IN-1k & IN-V2 & IN-R & IN-Sketch & IN-A & Avg. \\
\midrule
{Zero-shot} & 68.2 & 62.0 & 76.4 & 46.6 & 51.6 & 59.1 \\
\midrule
\multicolumn{7}{c}{\textbf{25 Shots}} \\
\midrule
{Full-FT} & 72.7 & 63.4 & 59.3 & 40.2 & 29.4 & 48.1 \\
{MovingAvg} & 72.9 & 63.5 & 61.6 & 41.1 & 29.6 & 48.9 \\
{Model Soups} & 74.5 & 65.8 & 62.7 & 43.5 & 35.0 & 51.8 \\
\midrule
{Linear Probing} & 72.0 & 63.2 & 72.4 & 43.8 & 47.6 & 56.7 \\
{LoRA} & 75.1 & 66.8 & 72.8 & 46.5 & 44.1 & 57.6 \\
{Random Mask} & 73.9 & 66.3 & 71.4 & 47.2 & 44.4 & 57.3 \\
{Mixout} & 74.2 & 66.3 & 66.2 & 44.7 & 39.7 & 54.2 \\
\rowcolor{gray!15}
\bottomrule
{\textbf{GMixout}} & 73.5 & 65.9 & 74.6 & 48.6 & 44.6 & 58.4 \\
\midrule
\multicolumn{7}{c}{\textbf{50 Shots}} \\
\midrule
{Full-FT} & 74.7 & 65.3 & 57.0 & 39.7 & 28.6 & 47.7 \\
{MovingAvg} & 74.7 & 63.4 & 76.1 & 47.1 & 51.6 & 59.6 \\
{Model Soups} & 76.5 & 67.4 & 62.1 & 44.0 & 34.4 & 52.0 \\
\midrule
{Linear Probing} & 73.3 & 63.8 & 71.2 & 43.3 & 46.5 & 56.2 \\
{LoRA} & 76.8 & 68.3 & 71.6 & 46.0 & 43.2 & 57.3 \\
{Random Mask} & 75.8 & 67.9 & 70.4 & 47.6 & 44.2 & 57.5 \\
{Mixout} & 76.0 & 67.2 & 65.1 & 44.8 & 39.3 & 54.1 \\
\rowcolor{gray!15}
\bottomrule
{\textbf{GMixout}} & 75.2 & 67.3 & 72.8 & 48.3 & 44.9 & 58.4 \\
\midrule
\multicolumn{7}{c}{\textbf{100 Shots}} \\
\midrule
{Full-FT} & 76.9 & 66.7 & 58.2 & 41.3 & 28.9 & 48.8 \\
{MovingAvg} & 76.9 & 63.3 & 76.0 & 46.8 & 51.2 & 59.3 \\
{Model Soups} & 78.4 & 69.0 & 62.8 & 45.5 & 35.3 & 53.1 \\
\midrule
{Linear Probing} & 74.7 & 64.6 & 70.3 & 43.2 & 45.6 & 55.9 \\
{LoRA} & 78.5 & 69.0 & 70.9 & 46.2 & 44.0 & 57.6 \\
{Random Mask} & 77.6 & 69.0 & 70.2 & 47.9 & 43.9 & 57.7 \\
{Mixout} & 77.6 & 68.5 & 65.3 & 45.4 & 38.0 & 54.3 \\
\rowcolor{gray!15}
\bottomrule
{\textbf{GMixout}} & 76.9 & 68.9 & 71.9 & 48.1 & 44.5 & 58.4 \\
\midrule
\multicolumn{7}{c}{\textbf{200 Shots}} \\
\midrule
{Full-FT} & 78.7 & 68.6 & 57.9 & 41.2 & 29.5 & 49.3 \\
{MovingAvg} & 78.7 & 63.4 & 75.5 & 46.2 & 50.8 & 59.0 \\
{Model Soups} & 80.2 & 70.5 & 63.3 & 46.0 & 35.7 & 53.9 \\
\midrule
{Linear Probing} & 75.8 & 65.6 & 70.0 & 43.4 & 45.9 & 56.2 \\
{LoRA} & 79.8 & 70.2 & 70.7 & 46.8 & 45.1 & 58.2 \\
{Random Mask} & 79.4 & 70.6 & 70.5 & 47.9 & 45.3 & 58.6 \\
{Mixout} & 79.1 & 69.9 & 64.4 & 45.0 & 36.9 & 54.0 \\
\rowcolor{gray!15}
\bottomrule
{\textbf{GMixout}} & 78.5 & 70.2 & 71.3 & 48.0 & 45.0 & 58.7 \\
\bottomrule

\end{tabular}
\end{adjustbox}
\label{tab:in_all_shots}
\end{table}

\end{document}

%% file: checklist.tex
\section*{NeurIPS Paper Checklist}

\begin{enumerate}

\item {\bf Claims}
    \item[] Question: Do the main claims made in the abstract and introduction accurately reflect the paper's contributions and scope?
    \item[] Answer: \answerYes{} 
    \item[] Justification: The implications of GMixout are illustrated in Figure~\ref{tab:main_1}, which appears in the introduction.
    \item[] Guidelines:
    \begin{itemize}
        \item The answer \answerNA{} means that the abstract and introduction do not include the claims made in the paper.
        \item The abstract and/or introduction should clearly state the claims made, including the contributions made in the paper and important assumptions and limitations. A \answerNo{} or \answerNA{} answer to this question will not be perceived well by the reviewers. 
        \item The claims made should match theoretical and experimental results, and reflect how much the results can be expected to generalize to other settings. 
        \item It is fine to include aspirational goals as motivation as long as it is clear that these goals are not attained by the paper. 
    \end{itemize}

\item {\bf Limitations}
    \item[] Question: Does the paper discuss the limitations of the work performed by the authors?
    \item[] Answer: \answerYes{} 
    \item[] Justification: The limitations of GMixout are discussed throughout the paper, including in the related work, proposed method, results sections.
    \item[] Guidelines:
    \begin{itemize}
        \item The answer \answerNA{} means that the paper has no limitation while the answer \answerNo{} means that the paper has limitations, but those are not discussed in the paper. 
        \item The authors are encouraged to create a separate ``Limitations'' section in their paper.
        \item The paper should point out any strong assumptions and how robust the results are to violations of these assumptions (e.g., independence assumptions, noiseless settings, model well-specification, asymptotic approximations only holding locally). The authors should reflect on how these assumptions might be violated in practice and what the implications would be.
        \item The authors should reflect on the scope of the claims made, e.g., if the approach was only tested on a few datasets or with a few runs. In general, empirical results often depend on implicit assumptions, which should be articulated.
        \item The authors should reflect on the factors that influence the performance of the approach. For example, a facial recognition algorithm may perform poorly when image resolution is low or images are taken in low lighting. Or a speech-to-text system might not be used reliably to provide closed captions for online lectures because it fails to handle technical jargon.
        \item The authors should discuss the computational efficiency of the proposed algorithms and how they scale with dataset size.
        \item If applicable, the authors should discuss possible limitations of their approach to address problems of privacy and fairness.
        \item While the authors might fear that complete honesty about limitations might be used by reviewers as grounds for rejection, a worse outcome might be that reviewers discover limitations that aren't acknowledged in the paper. The authors should use their best judgment and recognize that individual actions in favor of transparency play an important role in developing norms that preserve the integrity of the community. Reviewers will be specifically instructed to not penalize honesty concerning limitations.
    \end{itemize}

\item {\bf Theory assumptions and proofs}
    \item[] Question: For each theoretical result, does the paper provide the full set of assumptions and a complete (and correct) proof?
    \item[] Answer: \answerNA{} 
    \item[] Justification: Our work introduces a BVCL decomposition as a diagnostic tool, supported by empirical evidence rather than intended as a formal proof.
    \item[] Guidelines:
    \begin{itemize}
        \item The answer \answerNA{} means that the paper does not include theoretical results. 
        \item All the theorems, formulas, and proofs in the paper should be numbered and cross-referenced.
        \item All assumptions should be clearly stated or referenced in the statement of any theorems.
        \item The proofs can either appear in the main paper or the supplemental material, but if they appear in the supplemental material, the authors are encouraged to provide a short proof sketch to provide intuition. 
        \item Inversely, any informal proof provided in the core of the paper should be complemented by formal proofs provided in appendix or supplemental material.
        \item Theorems and Lemmas that the proof relies upon should be properly referenced. 
    \end{itemize}

    \item {\bf Experimental result reproducibility}
    \item[] Question: Does the paper fully disclose all the information needed to reproduce the main experimental results of the paper to the extent that it affects the main claims and/or conclusions of the paper (regardless of whether the code and data are provided or not)?
    \item[] Answer: \answerYes{} 
    \item[] Justification: Section~\ref{appx:impl}.
    \item[] Guidelines:
    \begin{itemize}
        \item The answer \answerNA{} means that the paper does not include experiments.
        \item If the paper includes experiments, a \answerNo{} answer to this question will not be perceived well by the reviewers: Making the paper reproducible is important, regardless of whether the code and data are provided or not.
        \item If the contribution is a dataset and\slash or model, the authors should describe the steps taken to make their results reproducible or verifiable. 
        \item Depending on the contribution, reproducibility can be accomplished in various ways. For example, if the contribution is a novel architecture, describing the architecture fully might suffice, or if the contribution is a specific model and empirical evaluation, it may be necessary to either make it possible for others to replicate the model with the same dataset, or provide access to the model. In general. releasing code and data is often one good way to accomplish this, but reproducibility can also be provided via detailed instructions for how to replicate the results, access to a hosted model (e.g., in the case of a large language model), releasing of a model checkpoint, or other means that are appropriate to the research performed.
        \item While NeurIPS does not require releasing code, the conference does require all submissions to provide some reasonable avenue for reproducibility, which may depend on the nature of the contribution. For example
        \begin{enumerate}
            \item If the contribution is primarily a new algorithm, the paper should make it clear how to reproduce that algorithm.
            \item If the contribution is primarily a new model architecture, the paper should describe the architecture clearly and fully.
            \item If the contribution is a new model (e.g., a large language model), then there should either be a way to access this model for reproducing the results or a way to reproduce the model (e.g., with an open-source dataset or instructions for how to construct the dataset).
            \item We recognize that reproducibility may be tricky in some cases, in which case authors are welcome to describe the particular way they provide for reproducibility. In the case of closed-source models, it may be that access to the model is limited in some way (e.g., to registered users), but it should be possible for other researchers to have some path to reproducing or verifying the results.
        \end{enumerate}
    \end{itemize}

\item {\bf Open access to data and code}
    \item[] Question: Does the paper provide open access to the data and code, with sufficient instructions to faithfully reproduce the main experimental results, as described in supplemental material?
    \item[] Answer: \answerYes{} 
    \item[] Justification: All of the datasets are publicly available (see Section~\ref{appx:datasets}). The source code is provided in the supplementary materials.
    \item[] Guidelines:
    \begin{itemize}
        \item The answer \answerNA{} means that paper does not include experiments requiring code.
        \item Please see the NeurIPS code and data submission guidelines (\url{https://neurips.cc/public/guides/CodeSubmissionPolicy}) for more details.
        \item While we encourage the release of code and data, we understand that this might not be possible, so \answerNo{} is an acceptable answer. Papers cannot be rejected simply for not including code, unless this is central to the contribution (e.g., for a new open-source benchmark).
        \item The instructions should contain the exact command and environment needed to run to reproduce the results. See the NeurIPS code and data submission guidelines (\url{https://neurips.cc/public/guides/CodeSubmissionPolicy}) for more details.
        \item The authors should provide instructions on data access and preparation, including how to access the raw data, preprocessed data, intermediate data, and generated data, etc.
        \item The authors should provide scripts to reproduce all experimental results for the new proposed method and baselines. If only a subset of experiments are reproducible, they should state which ones are omitted from the script and why.
        \item At submission time, to preserve anonymity, the authors should release anonymized versions (if applicable).
        \item Providing as much information as possible in supplemental material (appended to the paper) is recommended, but including URLs to data and code is permitted.
    \end{itemize}

\item {\bf Experimental setting/details}
    \item[] Question: Does the paper specify all the training and test details (e.g., data splits, hyperparameters, how they were chosen, type of optimizer) necessary to understand the results?
    \item[] Answer: \answerYes{} 
    \item[] Justification: Section~\ref{appx:impl} and Section~\ref{appx:baselines} and Section~\ref{main:results}.
    \item[] Guidelines:
    \begin{itemize}
        \item The answer \answerNA{} means that the paper does not include experiments.
        \item The experimental setting should be presented in the core of the paper to a level of detail that is necessary to appreciate the results and make sense of them.
        \item The full details can be provided either with the code, in appendix, or as supplemental material.
    \end{itemize}

\item {\bf Experiment statistical significance}
    \item[] Question: Does the paper report error bars suitably and correctly defined or other appropriate information about the statistical significance of the experiments?
    \item[] Answer: \answerYes{} 
    \item[] Justification: Section~\ref{appx:seeds} in the Appendix.
    \item[] Guidelines:
    \begin{itemize}
        \item The answer \answerNA{} means that the paper does not include experiments.
        \item The authors should answer \answerYes{} if the results are accompanied by error bars, confidence intervals, or statistical significance tests, at least for the experiments that support the main claims of the paper.
        \item The factors of variability that the error bars are capturing should be clearly stated (for example, train/test split, initialization, random drawing of some parameter, or overall run with given experimental conditions).
        \item The method for calculating the error bars should be explained (closed form formula, call to a library function, bootstrap, etc.)
        \item The assumptions made should be given (e.g., Normally distributed errors).
        \item It should be clear whether the error bar is the standard deviation or the standard error of the mean.
        \item It is OK to report 1-sigma error bars, but one should state it. The authors should preferably report a 2-sigma error bar than state that they have a 96\% CI, if the hypothesis of Normality of errors is not verified.
        \item For asymmetric distributions, the authors should be careful not to show in tables or figures symmetric error bars that would yield results that are out of range (e.g., negative error rates).
        \item If error bars are reported in tables or plots, the authors should explain in the text how they were calculated and reference the corresponding figures or tables in the text.
    \end{itemize}

\item {\bf Experiments compute resources}
    \item[] Question: For each experiment, does the paper provide sufficient information on the computer resources (type of compute workers, memory, time of execution) needed to reproduce the experiments?
    \item[] Answer: \answerYes{} 
    \item[] Justification: Section~\ref{appx:benchmark}.
    \item[] Guidelines:
    \begin{itemize}
        \item The answer \answerNA{} means that the paper does not include experiments.
        \item The paper should indicate the type of compute workers CPU or GPU, internal cluster, or cloud provider, including relevant memory and storage.
        \item The paper should provide the amount of compute required for each of the individual experimental runs as well as estimate the total compute. 
        \item The paper should disclose whether the full research project required more compute than the experiments reported in the paper (e.g., preliminary or failed experiments that didn't make it into the paper). 
    \end{itemize}
    
\item {\bf Code of ethics}
    \item[] Question: Does the research conducted in the paper conform, in every respect, with the NeurIPS Code of Ethics \url{https://neurips.cc/public/EthicsGuidelines}?
    \item[] Answer: \answerYes{} 
    \item[] Justification: Our work is based on public data.
    \item[] Guidelines:
    \begin{itemize}
        \item The answer \answerNA{} means that the authors have not reviewed the NeurIPS Code of Ethics.
        \item If the authors answer \answerNo, they should explain the special circumstances that require a deviation from the Code of Ethics.
        \item The authors should make sure to preserve anonymity (e.g., if there is a special consideration due to laws or regulations in their jurisdiction).
    \end{itemize}

\item {\bf Broader impacts}
    \item[] Question: Does the paper discuss both potential positive societal impacts and negative societal impacts of the work performed?
    \item[] Answer: \answerYes{} 
    \item[] Justification: Section~\ref{sec:impact}
    \item[] Guidelines:
    \begin{itemize}
        \item The answer \answerNA{} means that there is no societal impact of the work performed.
        \item If the authors answer \answerNA{} or \answerNo, they should explain why their work has no societal impact or why the paper does not address societal impact.
        \item Examples of negative societal impacts include potential malicious or unintended uses (e.g., disinformation, generating fake profiles, surveillance), fairness considerations (e.g., deployment of technologies that could make decisions that unfairly impact specific groups), privacy considerations, and security considerations.
        \item The conference expects that many papers will be foundational research and not tied to particular applications, let alone deployments. However, if there is a direct path to any negative applications, the authors should point it out. For example, it is legitimate to point out that an improvement in the quality of generative models could be used to generate Deepfakes for disinformation. On the other hand, it is not needed to point out that a generic algorithm for optimizing neural networks could enable people to train models that generate Deepfakes faster.
        \item The authors should consider possible harms that could arise when the technology is being used as intended and functioning correctly, harms that could arise when the technology is being used as intended but gives incorrect results, and harms following from (intentional or unintentional) misuse of the technology.
        \item If there are negative societal impacts, the authors could also discuss possible mitigation strategies (e.g., gated release of models, providing defenses in addition to attacks, mechanisms for monitoring misuse, mechanisms to monitor how a system learns from feedback over time, improving the efficiency and accessibility of ML).
    \end{itemize}
    
\item {\bf Safeguards}
    \item[] Question: Does the paper describe safeguards that have been put in place for responsible release of data or models that have a high risk for misuse (e.g., pre-trained language models, image generators, or scraped datasets)?
    \item[] Answer: \answerNA{} 
    \item[] Justification: We only use public datasets.
    \item[] Guidelines:
    \begin{itemize}
        \item The answer \answerNA{} means that the paper poses no such risks.
        \item Released models that have a high risk for misuse or dual-use should be released with necessary safeguards to allow for controlled use of the model, for example by requiring that users adhere to usage guidelines or restrictions to access the model or implementing safety filters. 
        \item Datasets that have been scraped from the Internet could pose safety risks. The authors should describe how they avoided releasing unsafe images.
        \item We recognize that providing effective safeguards is challenging, and many papers do not require this, but we encourage authors to take this into account and make a best faith effort.
    \end{itemize}

\item {\bf Licenses for existing assets}
    \item[] Question: Are the creators or original owners of assets (e.g., code, data, models), used in the paper, properly credited and are the license and terms of use explicitly mentioned and properly respected?
    \item[] Answer: \answerNA{} 
    \item[] Justification: The paper does not use existing assets.
    \item[] Guidelines:
    \begin{itemize}
        \item The answer \answerNA{} means that the paper does not use existing assets.
        \item The authors should cite the original paper that produced the code package or dataset.
        \item The authors should state which version of the asset is used and, if possible, include a URL.
        \item The name of the license (e.g., CC-BY 4.0) should be included for each asset.
        \item For scraped data from a particular source (e.g., website), the copyright and terms of service of that source should be provided.
        \item If assets are released, the license, copyright information, and terms of use in the package should be provided. For popular datasets, \url{paperswithcode.com/datasets} has curated licenses for some datasets. Their licensing guide can help determine the license of a dataset.
        \item For existing datasets that are re-packaged, both the original license and the license of the derived asset (if it has changed) should be provided.
        \item If this information is not available online, the authors are encouraged to reach out to the asset's creators.
    \end{itemize}

\item {\bf New assets}
    \item[] Question: Are new assets introduced in the paper well documented and is the documentation provided alongside the assets?
    \item[] Answer: \answerYes{} 
    \item[] Justification: The provided source code is reproducible and easy to follow.
    \item[] Guidelines:
    \begin{itemize}
        \item The answer \answerNA{} means that the paper does not release new assets.
        \item Researchers should communicate the details of the dataset\slash code\slash model as part of their submissions via structured templates. This includes details about training, license, limitations, etc. 
        \item The paper should discuss whether and how consent was obtained from people whose asset is used.
        \item At submission time, remember to anonymize your assets (if applicable). You can either create an anonymized URL or include an anonymized zip file.
    \end{itemize}

\item {\bf Crowdsourcing and research with human subjects}
    \item[] Question: For crowdsourcing experiments and research with human subjects, does the paper include the full text of instructions given to participants and screenshots, if applicable, as well as details about compensation (if any)? 
    \item[] Answer: \answerNA{} 
    \item[] Justification: The paper does not involve crowdsourcing nor research with human subjects.
    \item[] Guidelines:
    \begin{itemize}
        \item The answer \answerNA{} means that the paper does not involve crowdsourcing nor research with human subjects.
        \item Including this information in the supplemental material is fine, but if the main contribution of the paper involves human subjects, then as much detail as possible should be included in the main paper. 
        \item According to the NeurIPS Code of Ethics, workers involved in data collection, curation, or other labor should be paid at least the minimum wage in the country of the data collector. 
    \end{itemize}

\item {\bf Institutional review board (IRB) approvals or equivalent for research with human subjects}
    \item[] Question: Does the paper describe potential risks incurred by study participants, whether such risks were disclosed to the subjects, and whether Institutional Review Board (IRB) approvals (or an equivalent approval/review based on the requirements of your country or institution) were obtained?
    \item[] Answer: \answerNA{} 
    \item[] Justification: The paper does not involve crowdsourcing nor research with human subjects.
    \item[] Guidelines:
    \begin{itemize}
        \item The answer \answerNA{} means that the paper does not involve crowdsourcing nor research with human subjects.
        \item Depending on the country in which research is conducted, IRB approval (or equivalent) may be required for any human subjects research. If you obtained IRB approval, you should clearly state this in the paper. 
        \item We recognize that the procedures for this may vary significantly between institutions and locations, and we expect authors to adhere to the NeurIPS Code of Ethics and the guidelines for their institution. 
        \item For initial submissions, do not include any information that would break anonymity (if applicable), such as the institution conducting the review.
    \end{itemize}

\item {\bf Declaration of LLM usage}
    \item[] Question: Does the paper describe the usage of LLMs if it is an important, original, or non-standard component of the core methods in this research? Note that if the LLM is used only for writing, editing, or formatting purposes and does \emph{not} impact the core methodology, scientific rigor, or originality of the research, declaration is not required.
    \item[] Answer: \answerNA{} 
    \item[] Justification: We used LLM only for writing, editing, or formatting purposes.
    \item[] Guidelines:
    \begin{itemize}
        \item The answer \answerNA{} means that the core method development in this research does not involve LLMs as any important, original, or non-standard components.
        \item Please refer to our LLM policy in the NeurIPS handbook for what should or should not be described.
    \end{itemize}

\end{enumerate}